\documentclass[journal]{IEEEtran}
\IEEEoverridecommandlockouts
\usepackage{setspace}
\usepackage{cite}
\usepackage{amsmath,amssymb,amsfonts,mathtools,bm}
\usepackage{amsthm}
\usepackage{algorithm}
\usepackage{algpseudocode}
\usepackage{graphicx}
\usepackage{textcomp}
\usepackage{xcolor,float}
\usepackage{tabularx}
\usepackage{textcomp}
\usepackage{diagbox}
\usepackage{svg}
\usepackage{booktabs}
\usepackage{subfigure}
\usepackage{hyperref}

\newtheorem{problem}{Problem}
\usepackage{graphicx}
\usepackage{makecell}
\usepackage{amsthm}
\usepackage{caption}
\usepackage{sidecap}
\usepackage{microtype}
\usepackage{booktabs} 
\usepackage{multirow}
\usepackage{float}
\usepackage{adjustbox} 
\usepackage{amsmath}
\usepackage{amssymb}
\usepackage{colortbl}
\usepackage{makecell}
\usepackage{float}
\usepackage{url}
\usepackage{balance}
\usepackage{bbding}
\usepackage{hyperref}
\usepackage{graphicx}
\usepackage{sidecap}
\usepackage{microtype}
\usepackage{graphicx}
\usepackage{subfigure}
\usepackage{booktabs} 
\usepackage{multirow}
\usepackage{array}
\usepackage{subcaption}
\usepackage{graphicx}
\usepackage{makecell}
\usepackage{amsmath}
\usepackage{amssymb}

\usepackage{tikz}

\usepackage{changes} %宏包

\definecolor{lime}{HTML}{A6CE39}
\DeclareRobustCommand{\orcidicon}{%
    \begin{tikzpicture}
    \draw[lime, fill=lime] (0,0) 
    circle [radius=0.16] 
    node[white] {{\fontfamily{qag}\selectfont \tiny ID}};    \draw[white, fill=white] (-0.0625,0.095) 
    circle [radius=0.007];    \end{tikzpicture}
    \hspace{-2mm}}
\foreach \x in {A, ..., Z}{%
    \expandafter\xdef\csname orcid\x\endcsname{\noexpand\href{https://orcid.org/\csname orcidauthor\x\endcsname}{\noexpand\orcidicon}}
    }
\allowdisplaybreaks
\renewcommand{\baselinestretch}{1}

\begin{document}
% \doublespacing

\title{RadioDiff-$k^2$: Helmholtz Equation Informed Generative Diffusion Model for Multi-Path Aware Radio Map Construction}
\author{
Xiucheng Wang,\orcidA{} ~\IEEEmembership{Graduate Student Member,~IEEE,}
Qiming Zhang, \orcidB{}
Nan Cheng,\orcidC{}~\IEEEmembership{Senior Member,~IEEE,}
Ruijin Sun,\orcidD{}~\IEEEmembership{Member,~IEEE,}
Zan Li,\orcidE{}~\IEEEmembership{Fellow,~IEEE,}
Shuguang Cui,\orcidF{}~\IEEEmembership{Fellow,~IEEE,}\\
Xuemin (Sherman) Shen,\orcidG{}~\IEEEmembership{Fellow,~IEEE}

\thanks{ }% <-this % stops a space
\thanks{
\par This work was supported by the National Key Research and Development Program of China (2024YFB907500).
\par Xiucheng Wang, Nan Cheng, Ruijin Sun, and Zan Li are with the State Key Laboratory of ISN and School of Telecommunications Engineering, Xidian University, Xi’an 710071, China (e-mail: xcwang\_1@stu.xidian.edu.cn; dr.nan.cheng@ieee.org; \{sunruijin, zanli\}@xidian.edu.cn). \textit{Nan Cheng is the corresponding author}.
\par Qiming Zhang is with the School of Artificial Intelligence, Xidian University, Xi’an 710071, China (e-mail: 23009200991@stu.xidian.edu.cn);
\par Shuguang Cui is with the School of Science and Engineering (SSE), Shenzhen Future Network of Intelligence Institute (FNii-Shenzhen), and Guangdong Provincial Key Laboratory of Future Networks of Intelligence, The Chinese University of Hong Kong (Shenzhen), Shenzhen, China (e-mail: shuguangcui@cuhk.edu.cn);
\par Xuemin (Sherman) Shen is with the Department of Electrical and Computer Engineering, University of Waterloo, Waterloo, N2L 3G1, Canada (e-mail: sshen@uwaterloo.ca).
}

}
    
    \maketitle

\IEEEdisplaynontitleabstractindextext

\IEEEpeerreviewmaketitle

\begin{abstract}
In this paper, we propose a novel physics-informed generative learning approach, named RadioDiff-$k^2$, for accurate and efficient multipath-aware radio map (RM) construction. As future wireless communication evolves towards environment-aware paradigms, the accurate construction of RMs becomes crucial yet highly challenging. Conventional electromagnetic (EM)-based methods, such as full-wave solvers and ray-tracing approaches, exhibit substantial computational overhead and limited adaptability to dynamic scenarios. Although existing neural network (NN) approaches have efficient inferencing speed, they lack sufficient consideration of the underlying physics of EM wave propagation, limiting their effectiveness in accurately modeling critical EM singularities induced by complex multipath environments. To address these fundamental limitations, we propose a novel physics-inspired RM construction method guided explicitly by the Helmholtz equation, which inherently governs EM wave propagation. Specifically, based on the analysis of partial differential equations (PDEs), we theoretically establish a direct correspondence between EM singularities, which correspond to the critical spatial features influencing wireless propagation, and regions defined by negative wave numbers in the Helmholtz equation. We then design an innovative dual diffusion model (DM)-based large artificial intelligence framework comprising one DM dedicated to accurately inferring EM singularities and another DM responsible for reconstructing the complete RM using these singularities along with environmental contextual information. Experimental results demonstrate that the proposed RadioDiff-$k^2$ framework achieves state-of-the-art (SOTA) performance in both image-level RM construction and localization tasks, while maintaining inference latency within a few hundred milliseconds. Code is available at \url{https://github.com/UNIC-Lab/RadioDiff-k}.
\end{abstract}


\section{Introduction}
The paradigm shift in wireless communication from environment-independent operation to an environment-aware framework marks a fundamental transformation in the design and functionality of future 6G \cite{shen2023toward,6g}. This transition is driven by the increasing demand for intelligent, adaptive, and context-aware communication systems, particularly in the era of 6G networks, where network nodes must not only operate within but also dynamically interact with complex and time-varying environments \cite{zeng2021toward, Yang2023a}. A fundamental enabler of this transformation is the radio map (RM), which provides a spatial representation of critical wireless channel characteristics, including path loss, angle of arrival (AoA), and interference distribution \cite{zeng2024tutorial}. By integrating location-specific channel state information (CSI), RMs facilitate proactive network optimization, intelligent resource allocation, and real-time adaptation to environmental dynamics, offering a significant advantage over conventional reactive network management strategies \cite{YilmazBirkan2013}. One of the most impactful applications of RM lies in low-overhead channel estimation, where it enables infrastructure components, such as intelligent reflecting surfaces (IRS) and massive multiple-input multiple-output (MIMO) systems, to obtain high-fidelity CSI with minimal reliance on pilot signals \cite{wang2024tutorial}. This reduces training overhead, mitigates pilot contamination, and enhances spectral efficiency, making RM a cornerstone for sustainable and scalable 6G wireless communication. Furthermore, RM plays a pivotal role in the trajectory planning, coverage optimization, and interference mitigation of mobile network elements, such as autonomous aerial vehicles (AAVs) and satellites in space-air-ground integrated networks (SAGIN) \cite{cheng2019space,wang2022joint}. By precomputing the spatial variations of wireless channel properties, RM allows mobile nodes to anticipate and mitigate signal blockages, optimize handovers, and enhance connectivity resilience in highly dynamic environments.
\begin{figure}[t]
\captionsetup{font={small}, skip=16pt}
    \centering
    \vspace{-9pt}
    \subfigure[The RM of dominatent path.]
    {
       \centering
       \includegraphics[width=0.46\linewidth]{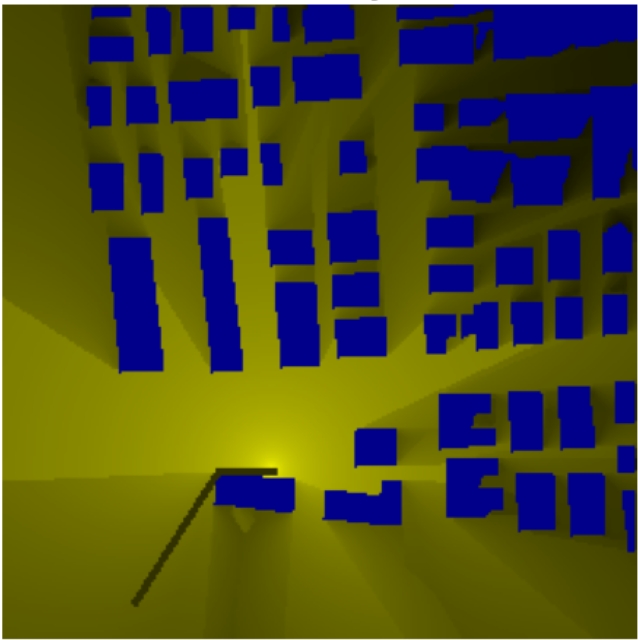}\label{fig-dpm}
    }
    \subfigure[The multi-path aware RM.]
    {
       \centering
       \includegraphics[width=0.46\linewidth]{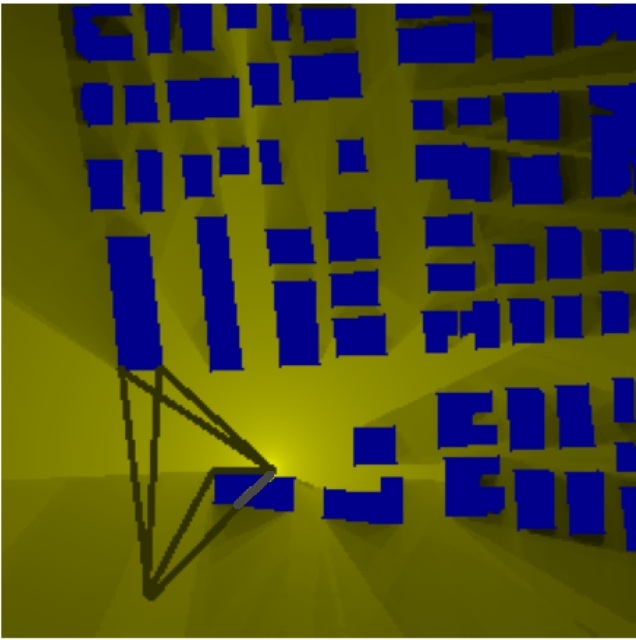}\label{fig-irt}
    }
    % \vspace{-10pt}
    \caption{Illustration of the RM, where (a) is the RM only considering the dominating path, (b) is the RM considering the multi-path effect, and the propagating path is shown for a specific location.}
    \label{fig-example}
    % \vspace{-12pt}
\end{figure}
\begin{figure*}[t]
\captionsetup{font={small}, skip=16pt}
    \centering
    \vspace{-9pt}
    \subfigure[RME-GAN \cite{zhang2023rme}.]
    {
       \centering
       \includegraphics[width=0.18\linewidth]{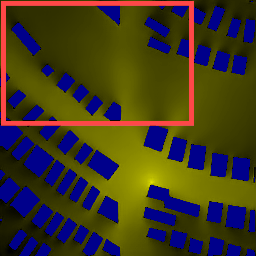}
    }
    \subfigure[UVM-Net \cite{zheng2024u}.]
    {
       \centering
       \includegraphics[width=0.18\linewidth]{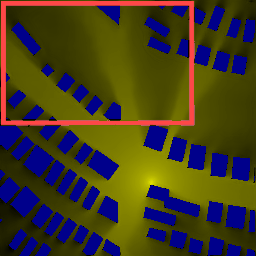}
    }
    \subfigure[RadioUNet \cite{levie2021RadioUNet}.]
    {
       \centering
       \includegraphics[width=0.18\linewidth]{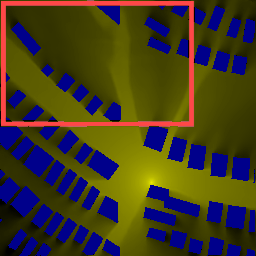}
    }
    \subfigure[RadioDiff \cite{wang2024radiodiff}.]
    {
       \centering
       \includegraphics[width=0.18\linewidth]{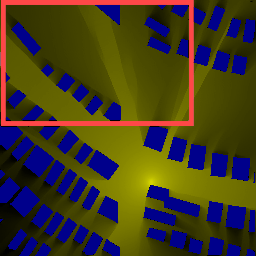}
    }
    \subfigure[Ground truth.]
    {
       \centering
       \includegraphics[width=0.18\linewidth]{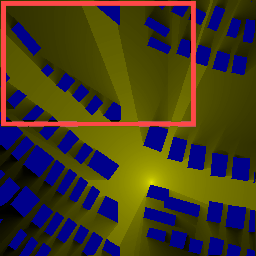}
    }
    \caption{Illustration of RMs constructed by various NN-based baseline methods, where RadioDiff is the SOTA NN-based RM construction method. The results indicate that existing NN-based approaches exhibit limited capability in capturing fine-grained texture features of the RM, particularly those associated with abrupt spatial variations in wireless channel characteristics. This deficiency hinders the ability of these methods to reliably detect electromagnetic singularities, thereby limiting their effectiveness in downstream wireless network optimization tasks.}
    \label{fig-baseline}
\end{figure*}
\begin{figure}[t]
    \centering
    \vspace{-9pt}
    \subfigure[The raw RM.]
    {
       \centering
       \includegraphics[width=0.46\linewidth]{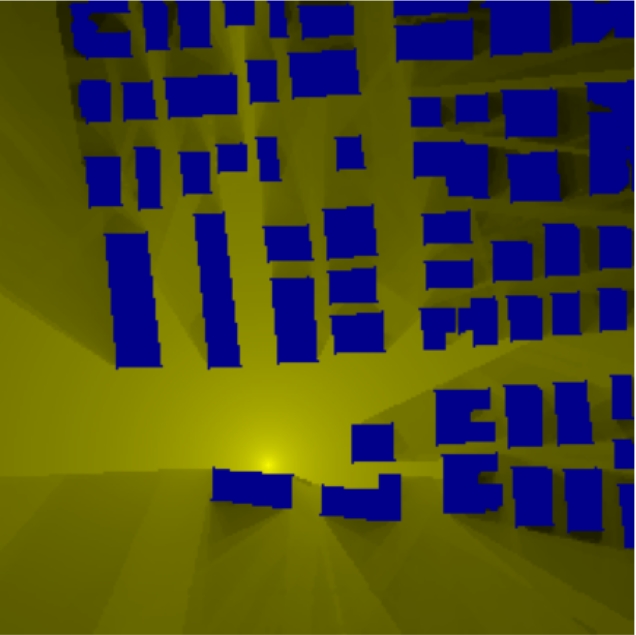}
    }
    \subfigure[The RM after the proposed Helmholtz equation informed filter.]
    {
       \centering
       \includegraphics[width=0.46\linewidth]{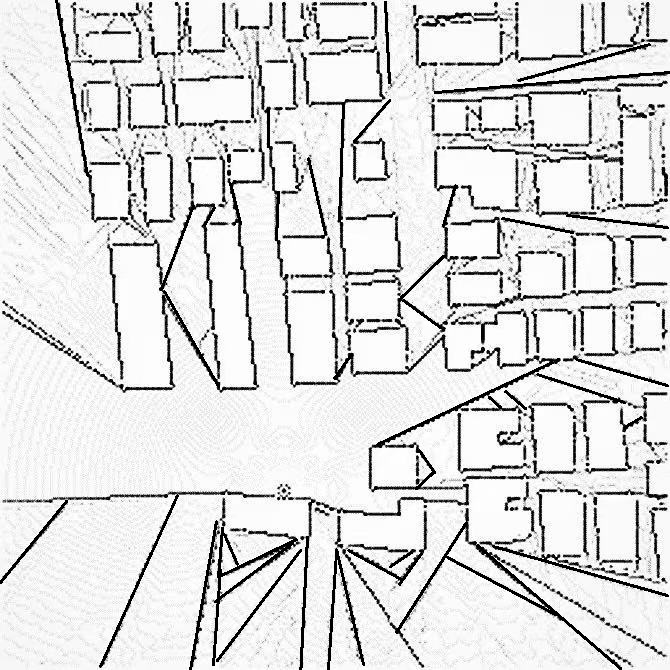}
    }
    \subfigure[The RM after Canny filter.]
    {
       \centering
       \includegraphics[width=0.46\linewidth]{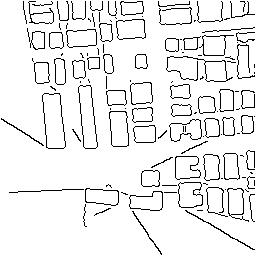}
    }
    \subfigure[The RM after LBP filter.]
    {
       \centering
       \includegraphics[width=0.46\linewidth]{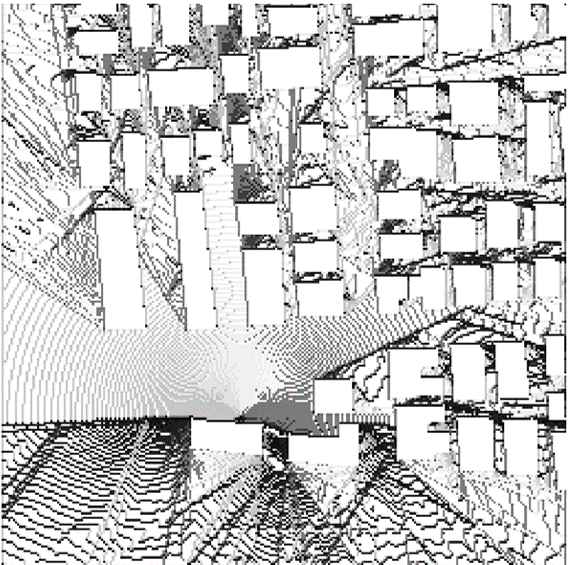}
    }
    % \vspace{-10pt}
    \caption{Illustration of the different texture extraction methods.}
    \label{fig-filter}
    \vspace{-10pt}
\end{figure}

However, the construction of RM is inherently challenging due to the simultaneous demands for efficiency and accuracy, particularly within highly dynamic 6G environments \cite{zeng2024tutorial}. Broadly, RM construction methods can be categorized into two distinct classes: electromagnetic (EM)-based high-precision methods \cite{oh2004mimo} and data-driven approaches emphasizing computational efficiency \cite{levie2021RadioUNet}. Traditional high-precision EM-based methods rely on solving Maxwell's equations via full-wave numerical simulations, providing highly accurate spatial distributions of wireless channel characteristics \cite{jones2013theory}. However, their computational complexity is substantial, typically requiring hours to simulate EM wave propagation even within meter-scale scenarios \cite{oh2004mimo}. To mitigate computational demands, approximate methods such as ray-tracing techniques have emerged, significantly reducing the required time to tens of minutes for kilometer-scale RM generation by simplifying EM wave propagation laws \cite{deschamps1972ray}. However, these EM-based approaches inherently assume a static environment during the computation period; this assumption is incompatible with the temporally and spatially dynamic characteristics of 6G networks, which include mobile network elements such as AAVs and satellites \cite{irt}. Consequently, recent attention has shifted toward neural network (NN)-based data-driven methods, offering efficiency through their rapid inference capabilities \cite{levie2021RadioUNet,zhang2023rme,wang2024radiodiff}. Nonetheless, existing NN-based approaches encounter significant limitations, notably their inability to accurately capture electromagnetic singularities resulting from intricate EM-environment interactions, as is shown in Fig.~\ref{fig-baseline}. These singularities, characterized by abrupt spatial variations in channel properties like pathloss, critically impact network performance by influencing factors such as user signal quality, AAV trajectory design, and beamforming alignment \cite{wang2022joint}. Moreover, traditional NN architectures, originally designed for tasks such as image processing or pattern recognition, inadequately address the physical nature and complexity of EM singularities, as these singularities are intrinsically tied to the physics of wave propagation rather than visual patterns or frequency domain characteristics \cite{levie2021RadioUNet}. An additional fundamental shortcoming of existing data-driven methods is their predominant focus on modeling the main propagation path, largely neglecting the complex yet critical multipath phenomena inherent in real-world wireless communication environments. Multipath propagation leads to numerous secondary signals arriving at the receiver from various directions and with diverse delays, significantly complicating the spatial distribution of EM singularities. Moreover, as shown in Fig.~\ref{fig-filter}, even classical digital filtering techniques fail to recover these fine-scale multipath-induced patterns \cite{jahne2005digital}, further underscoring the limitations of existing approaches. These observations highlight the urgent need for a new class of RM construction methods that not only infer the spatial distribution of channel characteristics but also incorporate a deep understanding of the physical mechanisms, particularly the multipath-driven singularities underpinning wave propagation.

However, achieving both efficiency and accuracy in RM construction remains challenging, motivating the development of novel methods to bridge the gap between the high precision of physics-based approaches and the computational efficiency of data-driven techniques. To address this challenge, this paper proposes a physics-inspired theoretical framework for effectively modeling the spatial distribution features of EM wave propagation. By explicitly incorporating the physics embedded in the EM wave equation, particularly the Helmholtz equation \cite{jones2013theory}, into the NN training process, our method enables the NN to capture spatial distribution characteristics intrinsic to EM wave propagation. Specifically, through rigorous theoretical analysis of the Helmholtz equation, we discover that electromagnetic singularities, which are key regions exhibiting abrupt changes in wireless channel features due to multipath effects, directly correspond to regions characterized by a complex wave number where $k^2<0$. Leveraging this novel insight, we devise a specialized NN architecture and training strategy that significantly enhances the NN's capability to identify and model abrupt variations in wireless channel characteristics caused by multipath propagation. Moreover, inspired by the success of the generative diffusion model (DM) \cite{ho2020denoising} for RM construction, where the DM-based method, such as RadioDiff \cite{wang2024radiodiff}, achieves much better performance than pixel-to-pixel methods, such as RadioUNet \cite{levie2021RadioUNet} and RME-GAN \cite{zhang2023rme}, within hundreds of milliseconds latency. The generative DM is used as the backbone to achieve high performance. Consequently, our physics-informed approach bridges the critical gap between traditional data-driven models and the underlying physical principles governing EM propagation, significantly advancing the accuracy and reliability of RM construction in complex 6G communication scenarios. The main contributions in this paper are summarized as follows.
\begin{enumerate}
    \item Theoretically, a physics-inspired RM construction theory is proposed in this paper, which explicitly integrates the underlying EM propagation characteristics governed by the Helmholtz equation into NN learning frameworks. Through an in-depth theoretical analysis of the Helmholtz equation, we reveal that electromagnetic singularities, corresponding to abrupt spatial variations in the wireless channel caused by multipath propagation, are characterized by regions where the $k^2<0$. This insight fundamentally enhances the ability of NNs to accurately capture and predict intricate spatial features, significantly improving the accuracy of RM-based wireless environmental awareness required by next-generation communication systems.
    \item Based on the physics-informed theory and partial differential equations (PDEs), we propose an innovative dual DM-based RM construction method specifically tailored for effectively modeling the complex multipath propagation environment. In our architecture, the first DM is explicitly designed and trained to predict the spatial distribution of EM singularities. The second NN leverages these learned singularities together with environmental context to accurately reconstruct the RM. This novel dual-stage DM approach ensures that intricate propagation characteristics induced by multipath effects are meticulously captured and effectively translated into the final RM.
    \item Different from traditional NN-based RM methods that primarily model the main path propagation features, our proposed method explicitly addresses multipath effects. By comprehensively incorporating multipath-induced singularities into RM construction, our method is particularly advantageous in highly dynamic and complex 6G scenarios.
    \item  Extensive experimental results demonstrate that the proposed method achieves remarkable improvements in widely adopted evaluation metrics, such as normalized mean square error (NMSE), root mean square error (RMSE), structural similarity (SSIM), and peak signal-to-noise ratio (PSNR). Moreover, it establishes a new SOTA baseline in downstream localization tasks, achieving sub-5-meter accuracy under both static and dynamic scenarios in heterogeneous line-of-the-sight (LoS) and non-LoS scenarios. Despite its dual-stage diffusion architecture, the model maintains practical inference latency within a few hundred milliseconds, facilitating its deployment in near-real-time applications.
\end{enumerate}

\section{Related Works and Preliminaries}
\subsection{Related Works}
RM construction methods can generally be classified into sampling-based and sampling-free approaches. Sampling-based methods rely on sparse pathloss measurements (SPM) collected at specific locations, which are then interpolated to estimate the RM over the entire area. These methods do not require prior knowledge of the environment or base station (BS) locations. Classic techniques include K-nearest neighbors (KNN) interpolation, which estimates unknown values by a weighted average of nearby measurements \cite{cover1967nearest}, and local multinomial regression, which fits a local linear model via least-squares minimization using nearby data points \cite{breidt2000local}. More sophisticated techniques, such as Kriging treat the interpolation as a stochastic process governed by spatial correlation, allowing for more accurate pathloss prediction by modeling the covariance structure of measurements \cite{qiu2024channel}.

Despite their simplicity, these methods suffer from two core limitations: a strong dependence on the availability of SPM and limited reconstruction accuracy, particularly in regions with sparse measurements or complex propagation environments \cite{levie2021RadioUNet,li2022radionet}. As a result, sampling-free RM construction has gained increasing attention. These approaches eliminate the need for measurement data in the target area and instead leverage environmental features—such as obstacle positions, heights, and BS coordinates—to infer location-specific channel information. Representative efforts include RadioUNet, which adapts the U-Net architecture from image-to-image translation tasks to RM generation using MSE-based supervision \cite{levie2021RadioUNet}, and RadioNet, which incorporates transformer-based attention mechanisms to capture global spatial dependencies in RM construction \cite{li2022radionet}. Further, graph neural networks (GNNs) have been applied to exploit the relational structure of spatial layouts for more expressive modeling \cite{chen2023graph}. While these methods offer improvements, most treat RM construction as a purely discriminative learning task and may struggle to capture the full distributional complexity of pathloss patterns in dynamic wireless environments. Some attempts, such as RME-GAN, introduced generative frameworks based on generative adversarial networks (GANs) to enhance output realism \cite{zhang2023rme}. However, RME-GAN remains dependent on sparse measurements, thus not truly sampling-free. In contrast, this work frames RM construction as a conditional generative modeling task, proposing a diffusion-based method that generates high-fidelity RMs solely from environmental and BS input features—without relying on sampling within the target region. This fundamentally shifts the RM construction paradigm, enabling stronger generalization and distribution modeling capabilities.
\begin{figure*}[!htbp]
\captionsetup{font={small}, skip=16pt}
    \centering
    \subfigure[The training process of the RadioDiff-$k^2$.]
    {
       \centering
       \includegraphics[width=0.85\linewidth]{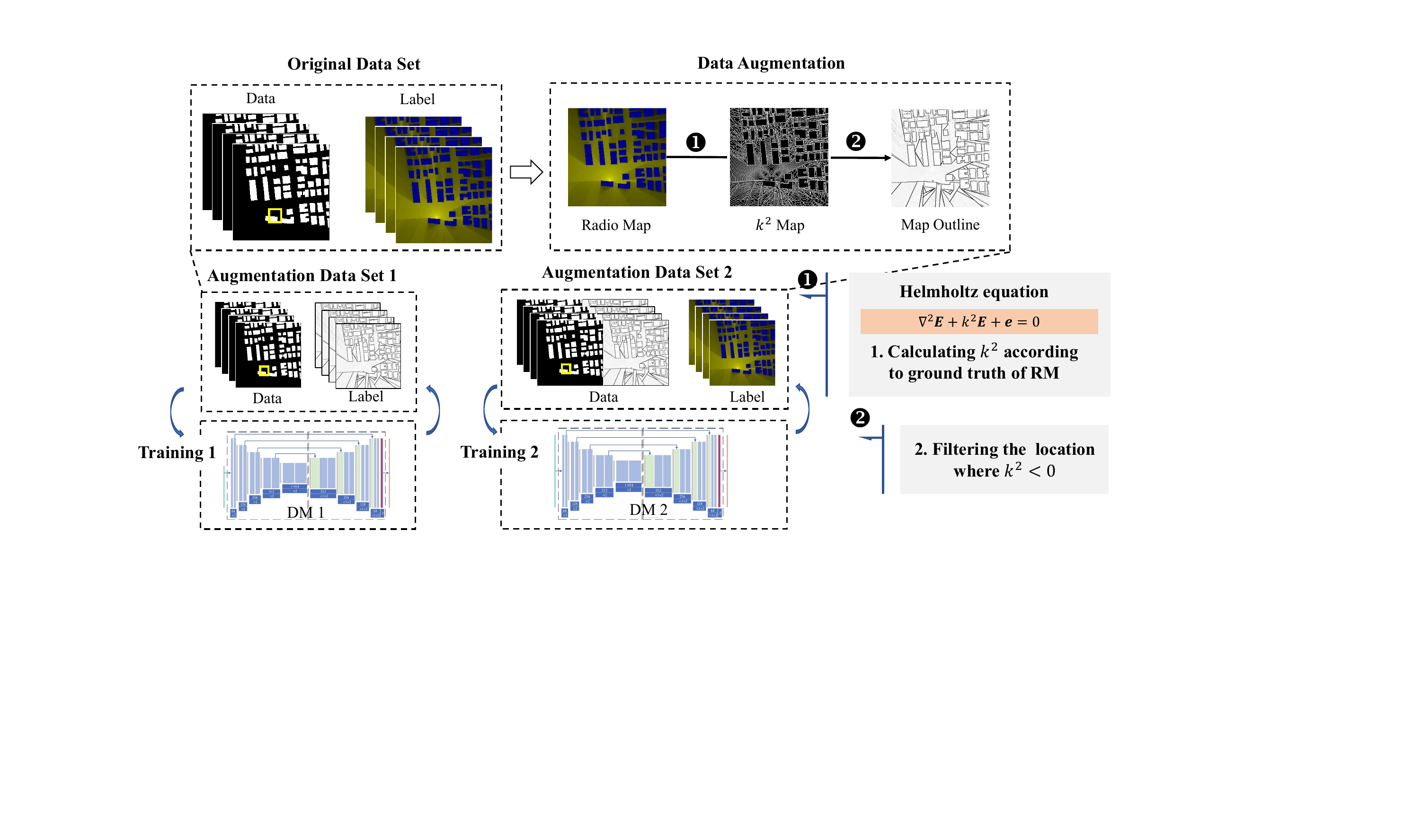}
    }
    \subfigure[The inferencing process of the RadioDiff-$k^2$.]
    {
       \centering
       \includegraphics[width=0.75\linewidth]{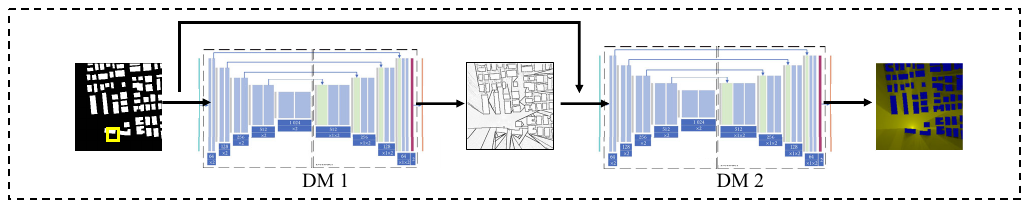}
    }
    \caption{Illustration of the proposed two-stage diffusion framework. (a) Training: The $k^2$ map is derived from the Helmholtz equation and GT, and points with $k^2<0$ form the Map Outline, refer to the Helmholtz-informed singularity map, used to train a dedicated DM. (b) Inference: Without solving the Helmholtz equation, the first DM predicts the Map Outline, which, together with environmental information, guides a second DM to generate the RM.}
    \vspace{-12pt}
    \label{fig-framework}
\end{figure*}

\subsection{Physics Informed Dual DM Framework}
\begin{algorithm}[h]
    \caption{Pipeline of RadioDiff-$k^2$}
    \label{alg-training}
    \begin{algorithmic}[1]
    \State \textbf{Input}: environment maps $\bm{H}_s$, dynamic maps $\bm{H}_d$, BS location $\bm{r}$, ground-truth RM $\bm{P}$, initialized DM1 for map outline and initialized DM2 for RM generation.
    \Statex
    \State \# \textbf{  Training Pipline  } \#
    \State \textbf{Stage 0: Data preprocessing}
    \State For each RM , compute $k^2_{\mathrm{eff}}(x)$ using~\eqref{eq-k-eff}.
    \State Define the outline mask $\bm{1}\!\left\{k^2_{\mathrm{eff}}(x) < 0\right\}$, which keep locations where $k^2_{\mathrm{eff}}$ is negative.
    \Statex

    \State \textbf{Stage 1: Train DM1 for map-outline generation}
    \State Use $(\bm{H}_s,\bm{H}_d,r)$ as inputs and train DM1 to predict the outline map.
    \Statex

    \State \textbf{Stage 2: Train DM2 for RM reconstruction}
    \State Use $(\bm{H}_s,\bm{H}_d,r)$, and the outline map generated by DM1 as inputs and train DM2 to generate the RM $\widehat{\bm{P}}$.
    \Statex
    \State \# \textbf{  Inferencing Pipline  } \#
    \State Input $(\bm{H}_s,\bm{H}_d,r)$ to the DM1 to get the outline map.
    \State Input $(\bm{H}_s,\bm{H}_d,r)$ and outline map generated by the DM1 to the DM2 to predict the RM.
    \end{algorithmic}
\end{algorithm}
\subsection{Score-Based Denoising Diffusion Model}
DMs have recently emerged as powerful generative frameworks capable of capturing complex data distributions and synthesizing high-fidelity samples \cite{ho2020denoising}. Among these, score-based diffusion models define a generative process grounded in stochastic differential equations (SDEs), wherein data undergoes progressive corruption and is later reconstructed by learning to approximate the gradient of the data log-likelihood, known as the score function \cite{song2020score}. Unlike classical DMs that adopt discrete Markov chains for the forward and reverse noise processes, score-based models perturb data in continuous time via SDEs, making them well-suited for solving inverse problems such as radio map reconstruction through Bayesian sampling principles. The forward process is formulated by the SDE as follows.
\begin{align}
d\bm{x} = f(\bm{x}, t) dt + g(t) d\bm{w},
\end{align}
where $f(\bm{x}, t)$ denotes the drift component, $g(t)$ is a time-dependent diffusion coefficient, and $d\bm{w}$ is a Wiener process. As time $t$ increases, the data distribution $p_t(\bm{x})$ transitions toward an isotropic Gaussian. Reconstructing the original data entails solving the reverse-time SDE as follows.
\begin{align}
d\bm{x} = \left[ f(\bm{x}, t) - g^2(t) \nabla_{\bm{x}} \log p_t(\bm{x}) \right] dt + g(t) d\bm{\bar{w}},
\end{align}
where $\nabla_{\bm{x}} \log p_t(\bm{x})$ is the score function, typically approximated by a neural network $s_\theta(\bm{x}, t)$. The objective is to train $s_\theta$ as follows.
\begin{align}
s_\theta(\bm{x}, t) \approx \nabla_{\bm{x}} \log p_t(\bm{x}).
\end{align}

For deterministic sampling, an equivalent probability flow ODE is derived:
\begin{align}
d\bm{x} = \left[ f(\bm{x}, t) - \frac{1}{2} g^2(t) \nabla_{\bm{x}} \log p_t(\bm{x}) \right] dt,
\end{align}
which allows sample generation without stochasticity, analogous to denoising diffusion probabilistic models (DDPMs). In DDPMs, the forward noise process is discretized as follows.
\begin{align}
q(\bm{x}_t | \bm{x}_{t-1}) = \mathcal{N}(\bm{x}_t; \alpha_t \bm{x}_{t-1}, \beta_t \bm{I}),
\end{align}
and the score is estimated via a learned denoiser $\epsilon_\theta$ as follows.
\begin{align}
s_\theta(\bm{x}, t) = -\frac{\epsilon_\theta(\bm{x}, t)}{\sqrt{1 - \bar{\alpha}_t}}.
\end{align}
This formulation highlights DDPM as a discrete approximation of continuous score-based models using a variance-preserving scheme.

In the context of RM construction, decoupled diffusion models (DDMs) have been proposed to improve generative stability, particularly in the RadioDiff framework \cite{huang2024decoupled}. Unlike standard models that inject noise directly, DDM first attenuates the original input to a zero baseline, followed by noise injection. The forward transition from $\bm{x}_0$ to $\bm{x}_t$ is governed as follows.
\begin{align}
q\left(\bm{x}_t \mid \bm{x}_0\right) = \mathcal{N}\left(\gamma_t \bm{x}_0, \delta_t^2 \bm{I}\right),
\end{align}
with time-dependent coefficients $\gamma_t$ and $\delta_t$ controlling signal decay and noise variance. The process is further described by the SDE as follows.
\begin{align}
&d \bm{x}_t = \bm{f}_t \bm{x}_t dt + g_t d\bm{\epsilon}_t,\\
&\bm{f}_t = \frac{d \log \gamma_t}{dt},\\
&\int_{0}^{t}\bm{f}_t dt = -x_0\\
&g_t^2 = \frac{d \delta_t^2}{dt} - 2 \bm{f}_t \delta_t^2,
\end{align}
where $\bm{f}_t$ dictates attenuation rate and $g_t^2$ captures noise accumulation over time.

The reverse-time generation process reconstructs $\bm{x}_0$ from $\bm{x}_t$ by solving the following equation.
\begin{align}
d \bm{x}_t = \left[\bm{f}_t \bm{x}_t - g_t^2 \nabla_{\bm{x}} \log q\left(\bm{x}_t\right)\right] dt + g_t d\overline{\bm{\epsilon}}_t.
\end{align}
Thanks to its two-phase design, DDM stabilizes training and improves sampling quality by decoupling data transformation and noise injection. The deterministic transformation toward the zero vector simplifies forward sampling as follows.
\begin{align}
q(\bm{x}_t|\bm{x}_0) = \mathcal{N}\left(\bm{x}_0 + \int_0^t \bm{f}_\tau d\tau, t\bm{I}\right),
\end{align}
and enables an efficient reverse update as follows.
\begin{align}
q\left(\bm{z}_{t-\Delta t} \mid \bm{z}_t, \bm{z}_0\right)  &=\mathcal{N}\left(\bm{z}_{t} +\int_t^{t-\Delta t} \bm{f}_t \mathrm{~d} t\right. \notag\\
& \left.\qquad\qquad-\frac{\Delta t}{\sqrt{t}} \bm{\epsilon}, \frac{\Delta t(t-\Delta t)}{t} \bm{I}\right).\label{ddm-reverse}
\end{align}
Through this structured perturbation mechanism, DDM enhances generation quality and computational efficiency, making it especially suitable for high-fidelity RM reconstruction in next-generation wireless networks.

\subsection{Helmholtz Wave Equation}
The derivation of the Helmholtz equation begins with Maxwell’s equations in a source-free, linear, isotropic, and time-invariant medium. Consider the time-harmonic forms of Maxwell’s equations with an $e^{j\omega t}$ dependence, where $\omega$ is the angular frequency. The two curl equations are as follows.
\begin{align}
\nabla \times \bm{E} = -j\omega\mu \bm{G}, \quad \nabla \times \bm{G} = j\omega\epsilon \bm{E},
\end{align}
where $\bm{E}$ and $\bm{G}$ denote the electric and magnetic fields, respectively, and $\epsilon$ and $\mu$ represent the permittivity and permeability of the medium.Taking the curl of Faraday’s law, $\nabla \times (\nabla \times \bm{E}) = -j\omega\mu \nabla \times \bm{G}$, and substituting Ampère’s law into it as follows.
\begin{align}
\nabla \times (\nabla \times \bm{E}) = -j\omega\mu (j\omega\epsilon \bm{E}) = -\omega^2 \mu\epsilon \bm{E}.
\end{align}
Utilizing the vector identity $\nabla \times (\nabla \times \bm{E}) = \nabla(\nabla \cdot \bm{E}) - \nabla^2 \bm{E}$, and assuming a source-free medium where $\nabla \cdot \bm{E} = 0$, the equation reduces to following form.
\begin{align}
\nabla^2 \bm{E} + \omega^2 \mu \epsilon \bm{E} = 0.
\end{align}
By defining the wavenumber $k = \omega \sqrt{\mu \epsilon}$, we obtain the vector Helmholtz equation as follows.
\begin{align}
\nabla^2 \bm{E} + k^2 \bm{E} = 0.\label{raw-helmholtz}
\end{align}
For fields with a radiation source, the Helmholtz equation can be expressed as follows.
\begin{align}
\nabla^2 \bm{E} + k^2 \bm{E} +\bm{e}= 0.
\end{align}
where $\bm{e}$ is the source term.

\section{System Model and Problem Formulation}
In this paper, we consider an RM construction scenario over a discrete spatial region modeled as an $N \times N$ grid. Each grid cell is assumed to be sufficiently small such that the pathloss within a cell remains approximately constant. Consequently, the RM can be represented by a matrix $\bm{P} \in \mathbb{R}^{N \times N}$, where each entry $P(i,j)$ denotes the pathloss value at the corresponding grid location. Similar to \cite{levie2021RadioUNet,wang2024radiodiff}, a single BS equipped with a dipole antenna serves as the sole radiation source within this region. The antenna emits EM energy as a spherical wave, and its position is specified by the tuple $r = \langle d_x, d_y, d_z \rangle$, where $d_z$ denotes the BS height and $(d_x, d_y)$ indicates its horizontal coordinates in the grid. Since the channel characteristics between different antennas in a massive MIMO system are largely independent, the single-antenna case generalizes to multi-antenna settings by simply varying the BS position across antennas \cite{jones2013theory}.

The environment comprises both static and dynamic obstacles. Static obstacles such as buildings are composed of homogeneous materials and exhibit consistent EM reflection and diffraction behavior. Following established assumptions \cite{zhang2023rme,levie2021RadioUNet,li2022radionet}, the interior of a static obstacle is modeled as a total EM shield, resulting in infinite pathloss, i.e., no signal propagation. The presence of static obstacles is described by a matrix $\bm{H}_s \in \mathbb{R}^{N \times N}$, where $H_s(i,j) = 0$ indicates the absence of a static obstacle at location $(i,j)$. In contrast, dynamic obstacles such as vehicles induce partial attenuation and scattering of EM waves due to their smaller physical size and lower elevation. Unlike static structures, dynamic objects do not completely block wave propagation. Their spatial distribution is encoded by the matrix $\bm{H}_d \in \mathbb{R}^{N \times N}$, with $H_d(i,j) = 0$ denoting the absence of dynamic obstacles at grid position $(i,j)$.

The goal of this work is to develop a NN $\bm{\mu}_\theta(\cdot)$, parameterized by $\bm{\theta}$, to predict the pathloss distribution $\hat{\bm{P}} \in \mathbb{R}^{N \times N}$ based on the environmental context and BS configuration. The network is trained to minimize the discrepancy between the predicted RM $\hat{\bm{P}}$ and the ground truth RM $\bm{P}$, quantified by a loss function $\mathcal{L}(\hat{\bm{P}}, \bm{P})$. The overall RM construction task can thus be formulated as the following optimization problem:
\begin{problem}\label{p1}
    \begin{align}
    &\min_{\bm{\theta}}&&\mathcal{L}(\hat{\bm{P}}, \bm{P}),\label{obj}\\
    &\text{s.t.} &&\hat{\bm{P}} = \bm{\mu}_{\bm{\theta}}(\bm{H}_s, \bm{H}_d, r)\tag{\ref{obj}a}.
    \end{align}
\end{problem}

\section{Helmholtz Equation Informed DM for RM Construction}
\subsection{Feature Analysis of Helmholtz Alignment}\label{sec-helmholtz-analysis}
To enable NNs to more effectively learn the spatial characteristics of EM-wave propagation, we ground the analysis in the time-harmonic Helmholtz equation. In a locally isotropic free-space region, the scalar field $\bm{u}(r)$\footnote{We define the scalar field $\bm{u}(r)$ to represent the time-harmonic electric field in a radially symmetric free-space region, which simplifies the Helmholtz equation into a one-dimensional radial form. This representation enables tractable analysis of wave attenuation behaviors—distinguishing between radiative $k^2 > 0$ and evanescent $k^2 < 0$ regimes—and supports the derivation of curvature-based indicators that guide singularity detection in radio maps.} satisfies
\begin{align}
&\nabla^2 \bm{u} + k^2 \bm{u} = 0,\\
&\nabla^2 \bm{u} = \frac{1}{r^2}\frac{d}{dr}\!\left(r^2\frac{d\bm{u}}{dr}\right)\notag\\
&\qquad= \frac{2}{r}\frac{d\bm{u}}{dr}+\frac{d^2\bm{u}}{dr^2}.
\end{align}
The resulting radial equation
\begin{align}
\frac{d^2 \bm{u}}{dr^2}+\frac{2}{r}\frac{d\bm{u}}{dr}+k^2 \bm{u}=0.
\label{eq:helmholtz-radial}
\end{align}
admits the outward spherical-wave solution $\bm{u}(r)=A e^{ikr}/r$.

Write $\bm{u}=Ae^{i\phi}$ with amplitude $A>0$ and phase $\phi$.
Expanding $\nabla^2(Ae^{i\phi})$ and separating real/imaginary parts as the eikonal–transport form \cite{jones2013theory}, gives
\begin{align}
\nabla^2 A - A\|\nabla\phi\|^2 + k^2 A = 0,\\
2\nabla A\!\cdot\!\nabla\phi + A\nabla^2\phi = 0.
\end{align}
Thus,
\begin{align}
&k^2=\|\nabla\phi\|^2-\frac{\nabla^2 A}{A}\\
&k_{\text{eff}}^2\;\triangleq\;-\frac{\nabla^2 A}{A}=k^2-\|\nabla\phi\|^2\le k^2.
\label{eq:amp-phase-identity}
\end{align}
If $k_{\text{eff}}^2(\bm{x})<0$, then $\|\nabla\phi(\bm{x})\|^2>k^2$, i.e., local spatial frequency/curvature dominates the medium wavenumber. This is a sufficient indicator for shadowing/diffraction/evanescent-like mutation zones in RM \cite{deschamps1972ray,balanis2016antenna}. Therefore, the field magnitude decays as $|\bm{u}(r)|\propto 1/r$, and the power obeys the inverse-square law, when the $k^2>0$
\begin{align}
I(r)\;\propto\;|\bm{u}(r)|^2=\frac{|A|^2}{r^2},
\end{align}
which is consistent with the Friis transmission relation in free space \cite{lee2012digital}.  In strongly confining or cutoff-like situations, the relevant component of the propagation constant becomes imaginary, giving rise to an evanescent behavior that can be captured by an effective imaginary wavenumber $k=i\gamma$, and $\gamma>0$, where $k^2<0$:
\begin{align}
\bm{u}(r)&=\frac{A e^{-\gamma r}}{r},\\
I(r)&\propto \frac{|A|^2 e^{-2\gamma r}}{r^2}.
\end{align}
Compared with the gradual $1/r$ decay of radiating waves, such evanescent regions exhibit rapid spatial attenuation and sharp texture transitions in radio maps. 

Motivated by this contrast, we introduce a local, curvature-based effective wavenumber
\begin{align}
k^2_{\text{eff}}(\bm{x})\;\triangleq\;-\frac{\nabla^2 \bm{u}(\bm{x})}{\bm{u}(\bm{x})},
\label{eq:keff}
\end{align}
which serves as a physics-informed indicator of high-variation zones: negative values of $k^2_{\text{eff}}$ are empirically aligned with abrupt changes due to shadowing, diffraction, or strong multipath. Unlike the physical $k^2=\omega^2\mu\varepsilon>0$ in homogeneous media, $k^2_{\text{eff}}$ is a data-derived quantity that summarizes local wavefront curvature and attenuation; it is therefore well suited to guide NNs toward EM-consistent discontinuities and singular structures that are often underrepresented by purely discriminative RM learners.

Given a RM to reprensetn the EM power $I=|u|^2$, we use $A=\sqrt{I}$ and compute the follows.
\begin{align}
    &k_{\log}(\bm{x})\triangleq -\nabla^2 \log\!\big(A(\bm{x})+\varepsilon\big),\\
    &k_{\text{eff}}^2(\bm{x})\triangleq -\frac{\nabla^2 A(\bm{x})}{A(\bm{x})+\varepsilon}.\label{eq-k-eff}
\end{align}
We primarily rely on the sign, such as $k_{\log}<0$, which is gain-invariant and robust under light smoothing and cross-scale persistence, and operationally actionable: negative belts co-locate with sharp power transitions and guide coverage enhancement, active sampling allocation, and high order stability tuning along user corridors. Importantly, we do not interpret  $k_{\text{eff}}^2$ as the material wavenumber; it is a curvature-based, physics-aligned prior for RMs.

\subsection{Discretization for Helmholtz-Aligned Indicators}
\begin{figure*}[!htbp]
\captionsetup{font={small}, skip=16pt}
    \centering
    \includegraphics[width=0.78\linewidth]{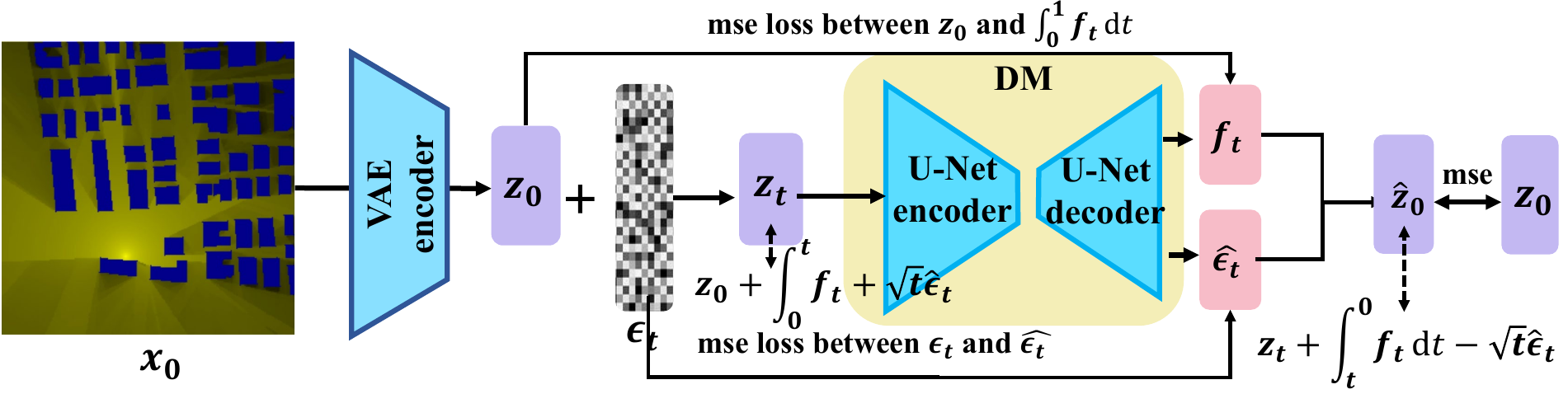}
    \caption{Illustration of the training details.}
    \vspace{-12pt}
    \label{fig-training}
\end{figure*}

All finite-difference operators are applied to $A=\sqrt{I}$ (or $\log(A+\varepsilon)$) rather than to the complex field:
\begin{align}
\frac{\partial^2 A}{\partial x^2}\bigg|_{(i,j)} \!\approx\! \frac{A(i{+}1,j)-2A(i,j)+A(i{-}1,j)}{h^2},\\
\frac{\partial^2 A}{\partial y^2}\bigg|_{(i,j)} \!\approx\! \frac{A(i,j{+}1)-2A(i,j)+A(i,j{-}1)}{h^2},
\end{align}
\begin{align}
    \nabla_h^2 A(i,j)\approx(&A(i{+}1,j)+A(i{-}1,j)+A(i,j{+}1)\\
    &+A(i,j{-}1)-4A(i,j))/h^2,
\end{align}
and the following.
\begin{align}
    &k_{\text{eff}}^2(i,j)= -\frac{\nabla_h^2 A(i,j)}{A(i,j)+\varepsilon},\\
    &k_{\log}(i,j)= -\nabla_h^2\log\!\big(A(i,j)+\varepsilon\big).
\end{align}

A single-scale 5-point Laplacian is $O(N)$ for $N$ grid points. With $s$ Gaussian scales, the cost is $O(sN)$ and remains negligible. We use mild Gaussian smoothing before differentiation and a small $\varepsilon$ to avoid division by near-zero $A$. Boundary conditions follow the RM acquisition setup, and anisotropic grids use $(h_x,h_y)$ accordingly. 

We refer to the raw EM power distribution, which is the RM, as the ``Field Strength Map", while the binary or grayscale mask extracted from the Helmholtz-informed curvature indicator $k_{eff}^2<0$ is denoted as the ``Map Outline". This outline highlights potential electromagnetic singularities and serves as an intermediate structural prior during training.

\subsection{Framework of the RadioDiff-$k^2$}
Building upon the theoretical analysis of the Helmholtz equation, we introduce Radiodiff-$k^2$, a novel dual-stage DM framework designed to address the complexities of RM construction, which is shown as Fig.~\ref{fig-framework}. This framework draws inspiration from existing SOTA RM generation architectures but incorporates significant enhancements to leverage the underlying physics of EM wave propagation for improved performance. The architecture consists of two integral components: a Variational Autoencoder (VAE) and a Denoising UNet. The VAE's primary role is to encode the radio map from image space into a latent hidden space, which allows the denoising UNet to operate more efficiently in this compressed representation. The denoising UNet, in turn, is responsible for predicting the $\epsilon_t$ and $\bm{f}_t$ terms required by the diffusion model during the denoising process with the condition of environment information and BS location, embedded by the cross attention, where the condition is used as the key and value vector of the cross attention \cite{wang2024radiodiff}. Moreover, the model adopts a conditional DM architecture, similar to the approach used in RadioDiff, enabling controllable RM generation based on various input conditions such as environmental features and the locations of buildings and base stations. Considering the drawbacks of directly enforcing a Helmholtz PDE loss inside the diffusion backbone, we deliberately avoid embedding PDE residuals into the training objective. In latent diffusion, the denoiser operates in a non-physical latent space where differential operators lack meaning; projecting a Helmholtz residual to image space \(x_0\) yields gradients that are typically much weaker than the diffusion objective and becomes ill-posed across noisy intermediates \(x_t\), while real-world RM measurements are power-only and phase-free, with uncertain boundaries and material parameters, further complicating faithful PDE enforcement \cite{ye2024diffusionedge}. Instead, we translate the physics prior into a measurement-faithful spatial feature: a Helmholtz-aligned curvature outline \(\mathcal{M}\) derived from the observable envelope \(A=\sqrt{I}\) via \(k_{\log}=-\nabla^{2}\log(A+\varepsilon)\), or \(k_{\mathrm{eff}}^{2}=-\nabla^{2}A/(A+\varepsilon))\), and we rely on its sign, such as \(k_{\log}<0)\) to obtain a gain-invariant, smoothing-robust indicator of shadow/diffraction belts. This outline is computed once in preprocessing, learned by DM1 as structured supervision, and injected into DM2 as conditional guidance through cross-attention together with environmental/BS inputs. The design preserves the efficiency of latent diffusion while delivering consistent gains over a single-stage diffusion baseline, as RadioDiff.

A key innovation of RadioDiff-$k^2$ lies in its integration of electromagnetic singularities, derived from the Helmholtz equation, to enhance RM generation. These singularities, corresponding to regions where $k^2 < 0$, are particularly important for accurately modeling areas of rapid change in wireless channel characteristics, such as sudden shifts in pathloss. To effectively capture these features, we designed a two-stage dual framework. In the first stage, a diffusion model is trained to learn and generate a feature map of regions where $k^2 < 0$, using environmental distributions and the base station location as input conditions. Once this model is trained, a second diffusion model is employed, which takes the predicted electric field singularity distribution map from the first model, along with the environmental features and base station location, to predict the final RM. It is important to note that, as illustrated in Fig.~\ref{fig-framework} and Algorithm~\ref{alg-training}, computations involving the Helmholtz equation are required only during the data preprocessing stage of training. In the inference stage, the framework merely performs two neural network forward passes. Since NN inference is highly efficient, the overall computational complexity is dominated solely by the network size.

This framework is inspired by curriculum learning \cite{wang2021survey}, where the complexity of the data distribution is tackled progressively. The first diffusion model focuses on capturing regions where wireless channel features undergo abrupt changes due to electromagnetic singularities. This reduces the complexity faced by the second model, which can then use the information on the distribution of the electric field singularities to generate a more accurate radio map. This stepwise decomposition of the RM generation problem improves both the efficiency and the effectiveness of the process, ensuring that each model specializes in a distinct aspect of the radio map’s spatial characteristics. Consequently, this method not only enhances the learning of fine-grained details but also provides better generalization and scalability in dynamic and complex environments.

\subsection{Training Method of RadioDiff-$k^2$}
In this work, the VAE is trained by minimizing the negative evidence lower bound (ELBO) of the RM, which consists of a reconstruction loss and a regularization term. Given an input sample $\bm{x}$, the encoder approximates the posterior distribution $q_\phi(\bm{z}|\bm{x})$ over latent variables $\bm{z}$, while the decoder seeks to reconstruct $\bm{x}$ from $\bm{z}$ via $p_\theta(\bm{x}|\bm{z})$. The overall objective is given by 
\begin{align}
    \mathcal{L}_{\mathrm{VAE}}(\theta, \phi; \bm{x}) = &\mathbb{E}_{q_\phi(\bm{z}|\bm{x})} \left[ -\log p_\theta(\bm{x}|\bm{z}) \right] \notag\\
    &+ D_{\mathrm{KL}}\left( q_\phi(\bm{z}|\bm{x})\ \|\ p(\bm{z}) \right).
\end{align}
The reconstruction loss is implemented as an $\ell_2$-norm between the input and output, assuming a Gaussian likelihood \cite{kingma2019introduction}. As for the denoising framework, Radiodiff-$k^2$ builds upon the DDM framework, where the inference process is governed by the reverse-time stochastic differential equation defined in \eqref{ddm-reverse}. This formulation necessitates the generative model to estimate two critical components: the deterministic drift term $\bm{f}_t$, which characterizes the latent trajectory of the denoising process, and the noise term $\hat{\bm{\epsilon}}_t$, which compensates for the stochastic perturbations introduced during the forward diffusion phase. In alignment with the design paradigm of RadioDiff, the ground truth for the drift term is defined as $\bm{f}_t = -\bm{z}_0$, where $\bm{z}_0$ denotes the clean latent representation sampled from the variational encoder. 

Accordingly, we define the drift loss as the MSE between the predicted drift term $\hat{\bm{f}}_t$ and its target:
\begin{align}
\mathcal{L}_{\text{drift}} = \mathbb{E}_t \left[ \left\| \hat{\bm{f}}_t + \bm{z}_0 \right\|_2^2 \right].
\end{align}
In parallel, the noise loss is constructed as the MSE between the predicted noise component $\hat{\bm{\epsilon}}_t$ and the actual injected noise:
\begin{align}
\mathcal{L}_{\text{noise}} = \mathbb{E}_t \left[ \left\| \hat{\bm{\epsilon}}_t - \bm{\epsilon}_t \right\|_2^2 \right].
\end{align}
Meanwhile, these losses ensure that the model captures both the deterministic and stochastic components of the generative process. Although this dual-objective training strategy fosters convergence to an accurate generative path, further analysis of the reverse-time formulation offers a critical enhancement. By setting the denoising step size $\Delta t = t$, the model can reconstruct the clean latent variable $\hat{\bm{z}}_0$ in a single step as follows.
\begin{align}
\hat{\bm{z}}_0 = \bm{z}_t + \int_0^t \hat{\bm{f}}_\tau d\tau - \sqrt{t}\,\hat{\bm{\epsilon}}_t.
\end{align}
This observation enables the incorporation of an auxiliary reconstruction loss, defined as follows.
\begin{align}
\mathcal{L}_{\text{recon}} = \left\| \hat{\bm{z}}_0 - \bm{z}_0 \right\|_2^2,
\end{align}
which provides a global constraint on the learned denoising trajectory. This auxiliary term enhances training stability and promotes more faithful recovery of the latent representation, particularly under constrained denoising budgets. Combining these components, the overall loss function guiding the optimization of Radiodiff-$k^2$ is given as follows.
\begin{align}
\mathcal{L}_{\text{total}} = \lambda_1 \mathcal{L}_{\text{drift}} + \lambda_2 \mathcal{L}_{\text{noise}} + \lambda_3 \mathcal{L}_{\text{recon}},
\end{align}
where $\lambda_1$, $\lambda_2$, and $\lambda_3$ are scalar weights that balance the influence of each component. These coefficients can be tuned based on empirical performance and task-specific constraints.

Furthermore, the continuous-time formulation of the decoupled diffusion framework grants Radiodiff-$k^2$ the ability to perform adaptive-step inference. By modifying $\Delta t$, the model dynamically adjusts the granularity of the denoising trajectory, offering a practical trade-off between generation fidelity and computational efficiency. In real-time applications with stringent latency requirements, coarse inference with fewer steps can be adopted, while more refined RM generation is achievable via finer denoising schedules. This operational flexibility ensures that Radiodiff-$k^2$ remains both accurate and efficient in dynamic wireless communication environments.

\section{Experimental Results}
\begin{table*}[t]
\captionsetup{font={small}, skip=16pt}
\centering
\caption{\textbf{Quantitative Comparison on DPM.} Results in bold red and underlined blue highlight the highest and second highest, respectively. The $\uparrow$ indicates metrics whereby higher values constitute improved outcomes, with higher values preferred for all other metrics. ``Rate w/ $k^2$ Map (\%)" denotes the percentage improvement of the proposed method when conditioned on the $k^2$ map versus the baseline RadioDiff without the $k^2$ map.}
\vspace{-12pt}
\resizebox{0.85\linewidth}{!}{
\begin{tabular}{@{}cc|ccccc|c@{}}
\toprule
\multicolumn{2}{c|}{Methods} &RME-GAN &RadioUNet & RadioDiff & RMDM  & RadioDiff-$k^2$ & Rate w/ $k^2$ Map (\%)  \\ \midrule
 & NMSE & 0.0096 & 0.0088 & 0.0072 & {\color[HTML]{00009B} \underline{0.0059}}  & {\color[HTML]{9A0000} \textbf{0.0043}} & {\color[HTML]{00adb5}$\downarrow$ 40.28\%}\\
 & RMSE & 0.0279 & 0.0266 & 0.0240 & {\color[HTML]{00009B} \underline{0.0214}}  & {\color[HTML]{9A0000} \textbf{0.0193}} & {\color[HTML]{00adb5}$\downarrow$ 19.58\%}\\
 & SSIM $\uparrow$ & 0.9431 & 0.9466 & 0.9560 & {\color[HTML]{00009B} \underline{0.9591}}  & {\color[HTML]{9A0000} \textbf{0.9773}} & {\color[HTML]{f08a5d}$\uparrow$ 2.232\%}\\ 
\multirow{-4}{*}{SRM} & PSNR $\uparrow$ & 31.35 & 31.77 & 32.67 & {\color[HTML]{00009B} \underline{33.67}}  & {\color[HTML]{9A0000} \textbf{34.46}} & {\color[HTML]{f08a5d}$\uparrow$ 5.483\%}\\ \midrule
 & NMSE & 0.0115 & 0.0107 & {\color[HTML]{00009B} \underline{0.0090}} & 0.0098  & {\color[HTML]{9A0000} \textbf{0.0054}} & {\color[HTML]{00adb5}$\downarrow$ 40.00\%}\\
 & RMSE & 0.0306 & 0.0291 & {\color[HTML]{00009B} \underline{0.0266}} & 0.0283  & {\color[HTML]{9A0000} \textbf{0.0208}} & {\color[HTML]{00adb5}$\downarrow$ 21.80\%}\\
 & SSIM $\uparrow$ & 0.9276 & 0.9291 & {\color[HTML]{00009B} \underline{0.9432}} & 0.9363  & {\color[HTML]{9A0000} \textbf{0.9704}} & {\color[HTML]{f08a5d}$\uparrow$ 2.884\%}\\ 
\multirow{-4}{*}{DRM}  & PSNR $\uparrow$ & 30.42 & 30.89 & {\color[HTML]{00009B} \underline{31.71}} & 31.13  & {\color[HTML]{9A0000} \textbf{33.79}} & {\color[HTML]{f08a5d}$\uparrow$ 6.569\%}\\ \bottomrule
\end{tabular}
}
% \vspace{-2mm}
\label{tab-dpm}
% \vspace{-12pt}
\end{table*}

\begin{table*}[t]
\captionsetup{font={small}, skip=16pt}
\centering
\caption{\textbf{Quantitative Comparison on IRT.} Results in bold red and underlined blue highlight the highest and second highest, respectively. ``Rate w/ $k^2$ Map (\%)" denotes the percentage improvement of the proposed method when conditioned on the $k^2$ map versus the baseline RadioDiff without the $k^2$ map.}
\vspace{-12pt}
\resizebox{0.85\linewidth}{!}{
\begin{tabular}{@{}cc|ccccc|c@{}}
\toprule
\multicolumn{2}{c|}{Methods} &RME-GAN &RadioUNet & RadioDiff & RMDM & RadioDiff-$k^2$ & Rate w/ $k^2$ Map (\%)  \\ \midrule
 & NMSE & 0.0155 & 0.0159 & 0.0121 & {\color[HTML]{00009B} \underline{0.0100}}  & {\color[HTML]{9A0000} \textbf{0.0066}} & {\color[HTML]{00adb5}$\downarrow$ 45.45\%}\\
 & RMSE & 0.0340 & 0.0344 & 0.0309  & {\color[HTML]{00009B} \underline{0.0278}}  & {\color[HTML]{9A0000} \textbf{0.0236}} & {\color[HTML]{00adb5}$\downarrow$ 23.62\%}\\
 & SSIM $\uparrow$ & 0.9123 & 0.9102 & {\color[HTML]{00009B} \underline{0.9268}} & 0.9180  & {\color[HTML]{9A0000} \textbf{0.9674}} & {\color[HTML]{f08a5d}$\uparrow$ 4.380\%}\\
\multirow{-4}{*}{MRM} & PSNR $\uparrow$ & 29.74 & 29.64 & 30.44  & {\color[HTML]{00009B} \underline{31.40}} & {\color[HTML]{9A0000} \textbf{32.68}} & {\color[HTML]{f08a5d}$\uparrow$ 7.352\%} \\  \bottomrule
\end{tabular}
}
\vspace{-12pt}
\label{tab:irt}
\end{table*}

% \begin{table*}[t]
% \centering
% \caption{\textbf{Quantitative Comparison on IRT.} Results in bold red and underlined blue highlight the highest and second highest, respectively.}
% \resizebox{0.8\linewidth}{!}{
% \begin{tabular}{@{}cc|cccccc@{}}
% \toprule
% \multicolumn{2}{c|}{Methods} &RME-GAN &RadioUNet & RadioDiff & DPM-GT & RadioDiff-k(Ours) & Rate(\%)  \\ \midrule
%  & NMSE & 0.0155 & 0.0159 & {\color[HTML]{00009B} \underline{0.0121}} & 0.0271 & {\color[HTML]{9A0000} \textbf{0.0066}} & {\color[HTML]{00adb5}$\downarrow$ 45.45\%}\\
%  & RMSE & 0.0340 & 0.0344 & {\color[HTML]{00009B} \underline{0.0309}} & 0.0464 & {\color[HTML]{9A0000} \textbf{0.0236}} & {\color[HTML]{00adb5}$\downarrow$ 23.62\%}\\
%  & SSIM $\uparrow$ & 0.9123 & 0.9102 & {\color[HTML]{00009B} \underline{0.9268}} & 0.8911 & {\color[HTML]{9A0000} \textbf{0.9674}} & {\color[HTML]{f08a5d}$\uparrow$ 4.380\%}\\
% \multirow{-4}{*}{IRT} & PSNR $\uparrow$ & 29.74 & 29.64 & {\color[HTML]{00009B} \underline{30.44}} & 26.90 & {\color[HTML]{9A0000} \textbf{32.68}} & {\color[HTML]{f08a5d}$\uparrow$ 7.352\%} \\  \bottomrule
% \end{tabular}
% }
% \label{tab:irt}
% \end{table*}

\subsection{Datasets and Metrics}
To rigorously assess the performance of the proposed method, we conduct experiments on three distinct datasets, each representing varying degrees of environmental complexity and EM propagation fidelity. These datasets are designed to evaluate the effectiveness of RM reconstruction under both static and dynamic wireless scenarios, including the presence or absence of multipath effects.  The foundation of our evaluation is the RadioMapSeer dataset, introduced in the Pathloss RM Construction Challenge \cite{yapar2023first}. It contains 700 uniquely structured urban maps, each incorporating a varying number of buildings derived from OpenStreetMap data across cities, urban areas, and villages. For each map, 80 transmitter positions are specified along with the corresponding pathloss ground truth (GT). All maps are converted into 256 × 256 binary morphological images, where each pixel denotes one square meter, with ‘1’ representing building interiors and ‘0’ denoting open areas. Transmitter and receiver heights are both set to 1.5 meters, with buildings uniformly modeled at 25 meters. The transmit power is fixed at 23 dBm, and the carrier frequency is 5.9 GHz. In particular, we consider three RM variants as follows. The GTs of these methods are all generated by electromagnetic ray tracing methods, so they can also be regarded as a performance comparison between neural network methods and computational electromagnetic methods.
\begin{itemize}
    \item Static Radio Maps (SRM), generated using the dominant path model (DPM) \cite{dpm} that only accounts for the main propagation path and the influence of large-scale static buildings;  
    \item Dynamic Radio Maps (DRM), also based on the dominant path model but additionally incorporating small-scale dynamic obstacles (e.g., vehicles) randomly distributed along roadways, thereby simulating temporal urban dynamics;  
    \item Multipath-Aware Radio Maps (MRM), produced using an intelligent ray tracing (IRT) \cite{irt} engine that includes multipath propagation with up to four environmental interactions per signal path. Due to computational constraints, the MRM dataset excludes dynamic obstacles and focuses solely on static structures.
\end{itemize}
To enhance the realism of our data, the DRM generation pipeline explicitly models dynamic reflectors and heterogeneous electromagnetic properties. Mobile objects such as vehicles are injected into scenes at traffic-relevant spatial and temporal scales, and different dielectric/conductive parameters are assigned to vehicles and to buildings so as to realistically reproduce vehicle-induced multipath fluctuations, partial occlusions, and localized attenuation differences. The training corpus comprises 600 distinct environments with 80 base-station placements per environment, while evaluation is performed on 100 disjoint, unseen environments with 80 base-station placements each. This setup constitutes a zero-shot, cross-environment generalization test that measures transferability to novel layouts and varying building densities.

Our framework is implemented in PyTorch and trained in two sequential stages using the AdamW optimizer with a cosine-decayed learning rate, initially set to $5 \times 10^{-5}$ and gradually annealed to $5 \times 10^{-6}$. In the first stage, we train a VAE on the radio map images from the full training split. The VAE employs a latent dimensionality of 128 with 3 input channels, and its training is conducted on four NVIDIA RTX Pro6000 GPUs with batch size of 6, requiring approximately 30 hours. The second stage trains the denoising U-Net component of the diffusion model using the fixed latent representations produced by the VAE. This step is performed on four NVIDIA RTX Pro6000 GPUs with a batch size of 64 and completes in around 24 hours. Throughout, all images are resized to $256 \times 256$ resolution. The diffusion model adopts a time horizon of $T = 1000$, uses a Gaussian starting distribution, and minimizes an $\ell_2$-based loss targeting the prediction of noise. Consistent with \cite{LDM}, the VAE encodes each image into a latent tensor of shape $64 \times 64 \times 4$, enabling efficient denoising in a compressed representation space.

To comprehensively evaluate the performance of RM reconstruction, we adopt a combination of classical error-based and perceptual quality metrics. Following previous works such as \cite{levie2021RadioUNet}, we employ normalized mean squared error (NMSE) and root mean squared error (RMSE) to quantify overall prediction accuracy. However, since RM reconstruction aims not only to minimize global error but also to preserve fine structural and textural features critical to communication performance, we further incorporate the structural similarity index measure (SSIM) and peak signal-to-noise ratio (PSNR) to assess perceptual and spatial fidelity.

NMSE measures the relative energy of the reconstruction error with respect to the ground truth, while RMSE reflects the square root of the averaged pixel-wise error. These metrics are defined as:
\begin{equation}
\text{NMSE} = \frac{\sum_{m=1}^M\sum_{n=1}^N (I_b(m,n) - I(m,n))^2}{\sum_{m=1}^M\sum_{n=1}^N I^2(m,n)},
\end{equation}
\begin{equation}
\text{RMSE} = \sqrt{\frac{1}{MN} \sum_{m=1}^{M} \sum_{n=1}^{N} \left( I_b(m,n) - I(m,n) \right)^2},
\end{equation}
where $I_b$ and $I$ denote the predicted and ground truth RM values, respectively, and $M$, $N$ are the image dimensions.

To better evaluate structural fidelity, we employ SSIM, which captures perceptual similarity by considering luminance, contrast, and structural correlation between the predicted and ground truth maps. SSIM is especially relevant in RM tasks due to the presence of high-frequency textures and edge features that reflect electromagnetic wave behavior. It is computed as:
\begin{equation}
\text{SSIM}(x, y) = \frac{(2\mu_x \mu_y + C_1)(2\sigma_{xy} + C_2)}{(\mu_x^2 + \mu_y^2 + C_1)(\sigma_x^2 + \sigma_y^2 + C_2)},
\end{equation}
where $\mu_x, \mu_y$ are the local means, $\sigma_x, \sigma_y$ are variances, and $\sigma_{xy}$ is the covariance between inputs $x$ and $y$. Constants $C_1$ and $C_2$ stabilize the division and are derived from the dynamic range $L$ of the image.

Lastly, we include PSNR to measure the reconstruction fidelity relative to the maximum possible signal intensity, with a particular focus on edge preservation, which is crucial in reflecting signal boundaries in RMs. It is defined as:
\begin{equation}
\text{PSNR} = 10 \log_{10} \left( \frac{r^2}{\text{MSE}} \right),
\end{equation}
where $r$ denotes the maximum pixel value in the image. Higher PSNR values correspond to lower reconstruction error and better visual quality. These four metrics provide a well-rounded evaluation framework, enabling both quantitative accuracy and perceptual quality assessment of radio map reconstruction performance.

\begin{figure*}
\captionsetup{font={small}, skip=16pt}
\centering
\begin{adjustbox}{valign=t}
\begin{tabular}{ccccccccc}
\includegraphics[width=0.14\linewidth]{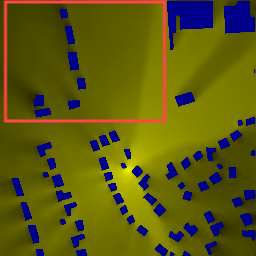} \hspace{-4mm} &
\includegraphics[width=0.14\linewidth]{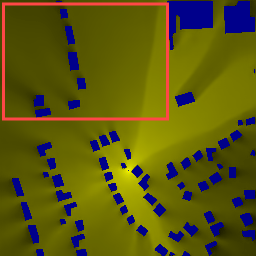}   \hspace{-4mm} &
\includegraphics[width=0.14\linewidth]{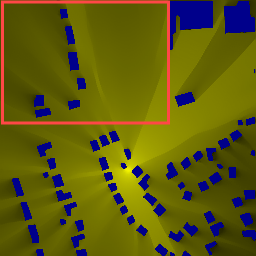}  \hspace{-4mm} &
\includegraphics[width=0.14\linewidth]{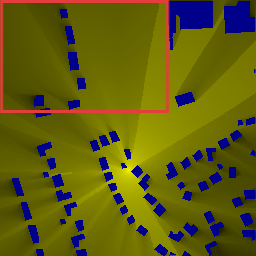}  \hspace{-4mm} &
\includegraphics[width=0.14\linewidth]{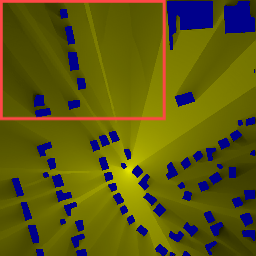}  \hspace{-4mm} &
\includegraphics[width=0.14\linewidth]{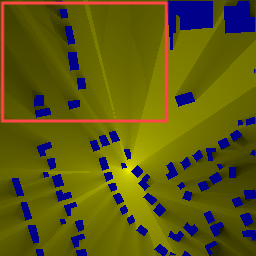}  \hspace{-4mm} &
\end{tabular}
\end{adjustbox}
\vspace{0.1mm}

\begin{adjustbox}{valign=t}
\begin{tabular}{ccccccccc}
\includegraphics[width=0.14\linewidth]{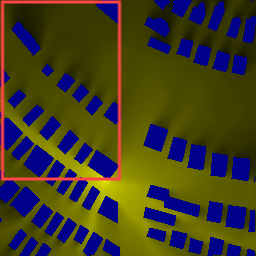} \hspace{-4mm} &
\includegraphics[width=0.14\linewidth]{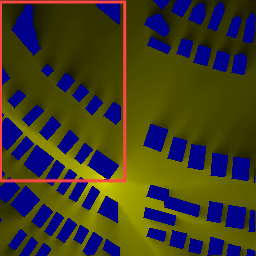}   \hspace{-4mm} &
\includegraphics[width=0.14\linewidth]{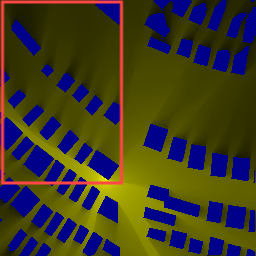}  \hspace{-4mm} &
\includegraphics[width=0.14\linewidth]{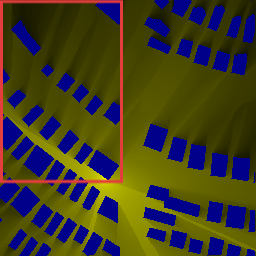}  \hspace{-4mm} &
\includegraphics[width=0.14\linewidth]{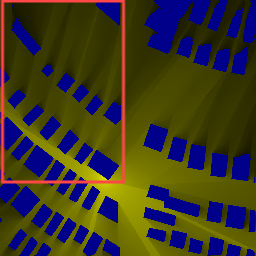}  \hspace{-4mm} &
\includegraphics[width=0.14\linewidth]{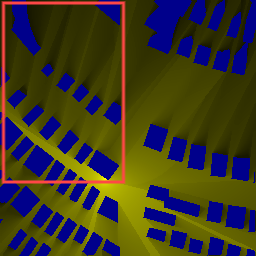}  \hspace{-4mm} &
\end{tabular}
\end{adjustbox}
\vspace{0.1mm}

\begin{adjustbox}{valign=t}
\begin{tabular}{ccccccccc}
\includegraphics[width=0.14\linewidth]{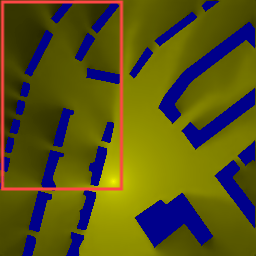} \hspace{-4mm} &
\includegraphics[width=0.14\linewidth]{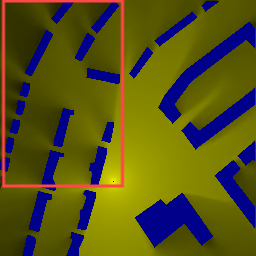}   \hspace{-4mm} &
\includegraphics[width=0.14\linewidth]{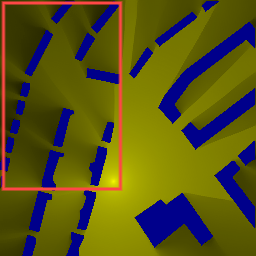}  \hspace{-4mm} &
\includegraphics[width=0.14\linewidth]{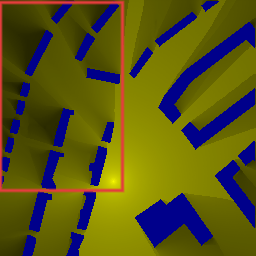}  \hspace{-4mm} &
\includegraphics[width=0.14\linewidth]{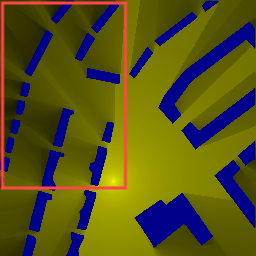}  \hspace{-4mm} &
\includegraphics[width=0.14\linewidth]{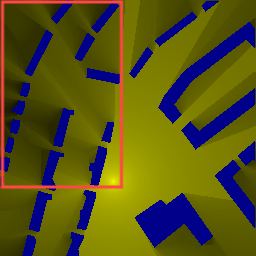}  \hspace{-4mm} &
\end{tabular}
\end{adjustbox}
\vspace{0.1mm}

\begin{adjustbox}{valign=t}
\begin{tabular}{ccccccccc}
\includegraphics[width=0.14\linewidth]{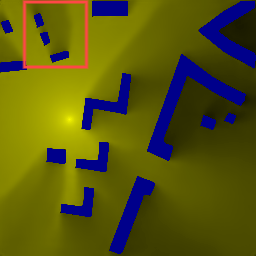} \hspace{-4mm} &
\includegraphics[width=0.14\linewidth]{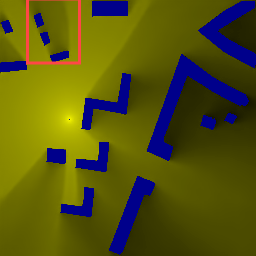}   \hspace{-4mm} &
\includegraphics[width=0.14\linewidth]{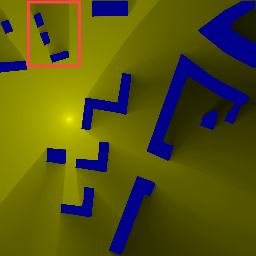}  \hspace{-4mm} &
\includegraphics[width=0.14\linewidth]{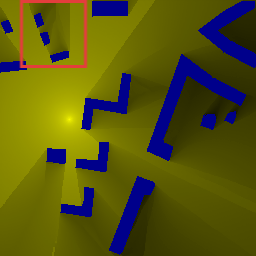}  \hspace{-4mm} &
\includegraphics[width=0.14\linewidth]{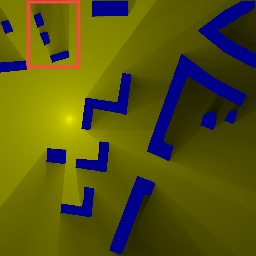}  \hspace{-4mm} &
\includegraphics[width=0.14\linewidth]{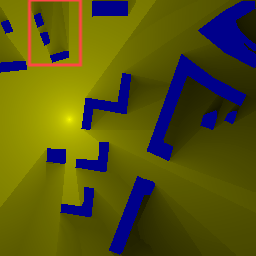}  \hspace{-4mm} &

\\
RME-GAN \hspace{-4mm} &
RadioUNet \hspace{-4mm} &
RadioDiff \hspace{-4mm} &
RMDM \hspace{-4mm} &
RadioDiff-$k^2$ \hspace{-4mm} &
Ground Truth
\\

\end{tabular}
\end{adjustbox}
\vspace{-12pt}
\caption{The comparisons of constructed SRM on different methods.}
\vspace{-12pt}
\label{fig-srm}
% \vspace{0.1mm}
\end{figure*}

\begin{figure*}[ht]
\captionsetup{font={small}, skip=16pt}
\centering
\begin{adjustbox}{valign=t}
\begin{tabular}{ccccccccc}
\includegraphics[width=0.14\linewidth]{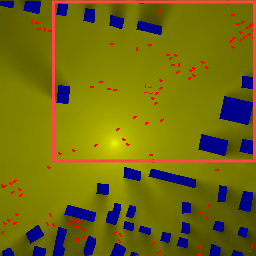} \hspace{-4mm} &
\includegraphics[width=0.14\linewidth]{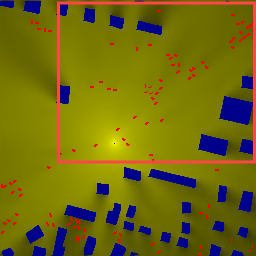}   \hspace{-4mm} &
\includegraphics[width=0.14\linewidth]{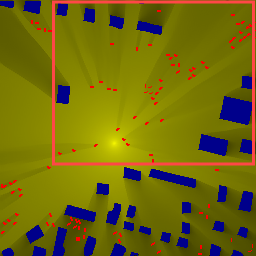}  \hspace{-4mm} &
\includegraphics[width=0.14\linewidth]{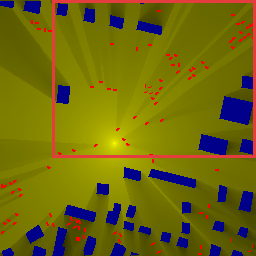}  \hspace{-4mm} &
\includegraphics[width=0.14\linewidth]{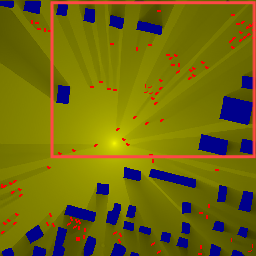}  \hspace{-4mm} &
\includegraphics[width=0.14\linewidth]{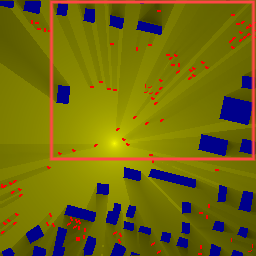}  \hspace{-4mm} &
\end{tabular}
\end{adjustbox}
\vspace{0.1mm}

\begin{adjustbox}{valign=t}
\begin{tabular}{ccccccccc}
\includegraphics[width=0.14\linewidth]{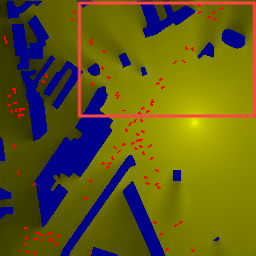} \hspace{-4mm} &
\includegraphics[width=0.14\linewidth]{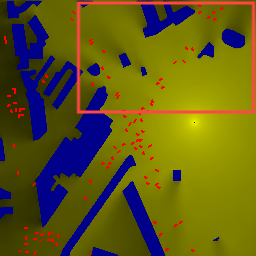}   \hspace{-4mm} &
\includegraphics[width=0.14\linewidth]{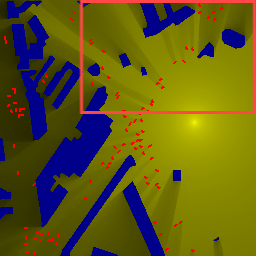}  \hspace{-4mm} &
\includegraphics[width=0.14\linewidth]{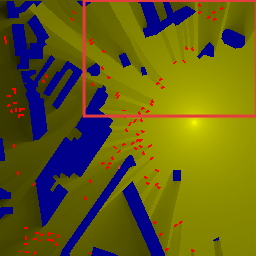}  \hspace{-4mm} &
\includegraphics[width=0.14\linewidth]{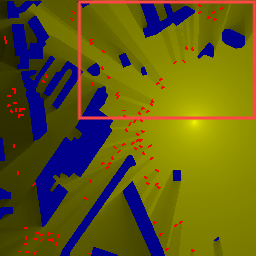}  \hspace{-4mm} &
\includegraphics[width=0.14\linewidth]{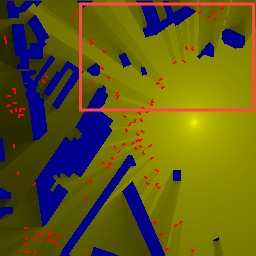}  \hspace{-4mm} &
\end{tabular}
\end{adjustbox}
\vspace{0.1mm}

\begin{adjustbox}{valign=t}
\begin{tabular}{ccccccccc}
\includegraphics[width=0.14\linewidth]{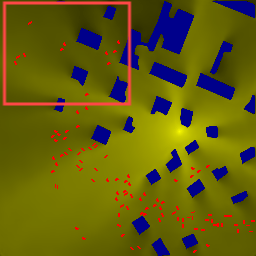} \hspace{-4mm} &
\includegraphics[width=0.14\linewidth]{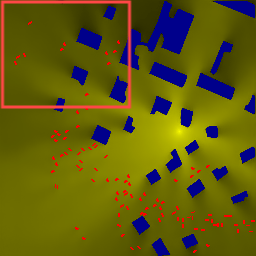}   \hspace{-4mm} &
\includegraphics[width=0.14\linewidth]{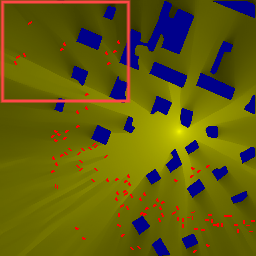}  \hspace{-4mm} &
\includegraphics[width=0.14\linewidth]{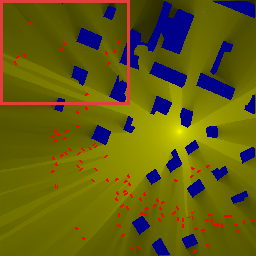}  \hspace{-4mm} &
\includegraphics[width=0.14\linewidth]{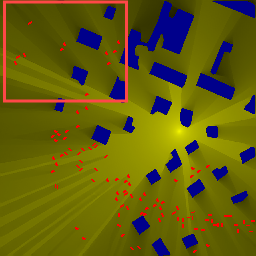}  \hspace{-4mm} &
\includegraphics[width=0.14\linewidth]{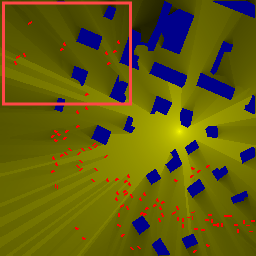}  \hspace{-4mm} &
\end{tabular}
\end{adjustbox}
\vspace{0.1mm}

\begin{adjustbox}{valign=t}
\begin{tabular}{ccccccccc}
\includegraphics[width=0.14\linewidth]{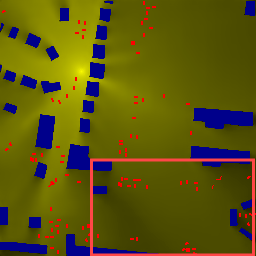} \hspace{-4mm} &
\includegraphics[width=0.14\linewidth]{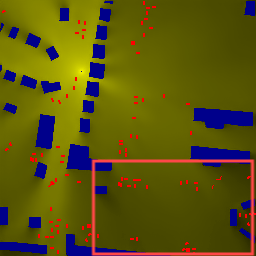}   \hspace{-4mm} &
\includegraphics[width=0.14\linewidth]{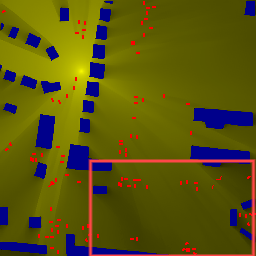}  \hspace{-4mm} &
\includegraphics[width=0.14\linewidth]{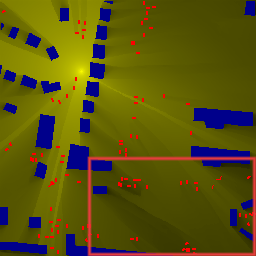}  \hspace{-4mm} &
\includegraphics[width=0.14\linewidth]{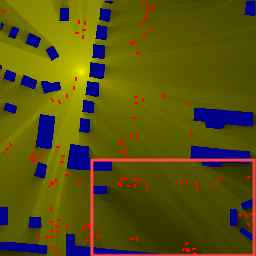}  \hspace{-4mm} &
\includegraphics[width=0.14\linewidth]{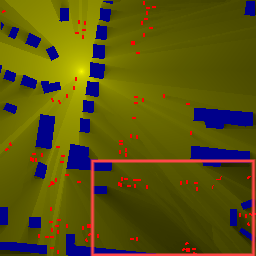}  \hspace{-4mm} &

\\
RME-GAN \hspace{-4mm} &
RadioUNet \hspace{-4mm} &
RadioDiff \hspace{-4mm} &
RMDM \hspace{-4mm} &
RadioDiff-$k^2$ \hspace{-4mm} &
Ground Truth
\\

\end{tabular}
\end{adjustbox}
\vspace{-8pt}
\caption{The comparisons of constructed DRM on different methods. The buildings are colored blue, while the cars are colored red.}
\vspace{-13pt}
\label{fig-drm}
% \vspace{0.1mm}
\end{figure*}

\begin{figure*}[ht]
\captionsetup{font={small}, skip=16pt}
\centering
\begin{adjustbox}{valign=t}
\begin{tabular}{ccccccccc}
\includegraphics[width=0.14\linewidth]{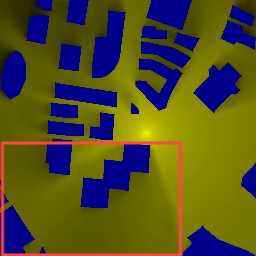} \hspace{-4mm} &
\includegraphics[width=0.14\linewidth]{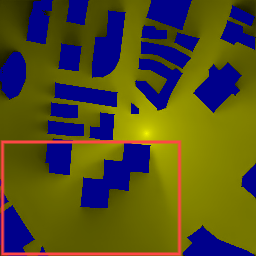}   \hspace{-4mm} &
\includegraphics[width=0.14\linewidth]{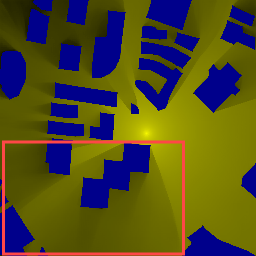}  \hspace{-4mm} &
\includegraphics[width=0.14\linewidth]{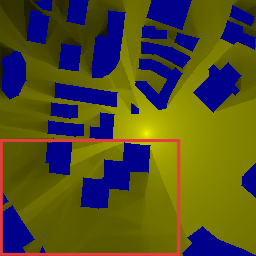}  \hspace{-4mm} &
\includegraphics[width=0.14\linewidth]{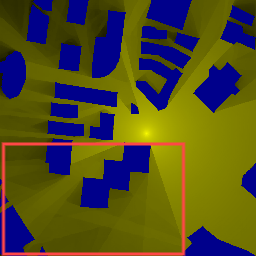}  \hspace{-4mm} &
\includegraphics[width=0.14\linewidth]{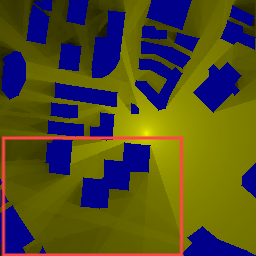}  \hspace{-4mm} &
\end{tabular}
\end{adjustbox}
\vspace{0.1mm}

\begin{adjustbox}{valign=t}
\begin{tabular}{ccccccccc}
\includegraphics[width=0.14\linewidth]{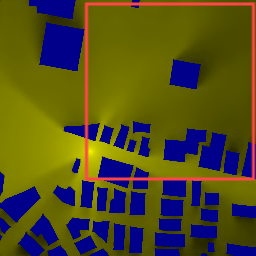} \hspace{-4mm} &
\includegraphics[width=0.14\linewidth]{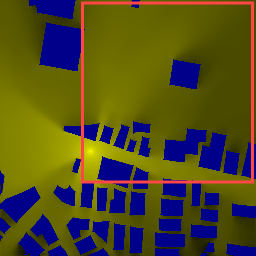}   \hspace{-4mm} &
\includegraphics[width=0.14\linewidth]{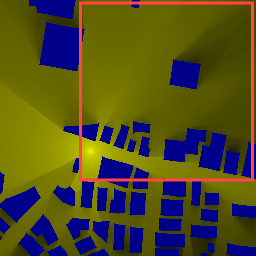}  \hspace{-4mm} &
\includegraphics[width=0.14\linewidth]{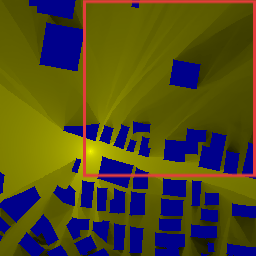}  \hspace{-4mm} &
\includegraphics[width=0.14\linewidth]{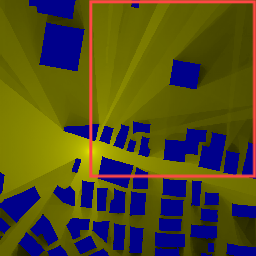}  \hspace{-4mm} &
\includegraphics[width=0.14\linewidth]{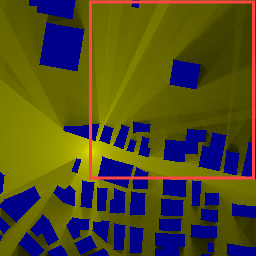}  \hspace{-4mm} &
\end{tabular}
\end{adjustbox}
\vspace{0.1mm}

\begin{adjustbox}{valign=t}
\begin{tabular}{ccccccccc}
\includegraphics[width=0.14\linewidth]{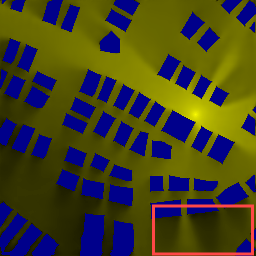} \hspace{-4mm} &
\includegraphics[width=0.14\linewidth]{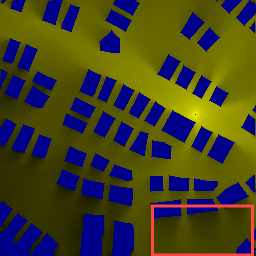}   \hspace{-4mm} &
\includegraphics[width=0.14\linewidth]{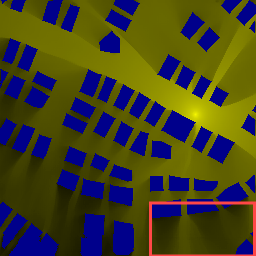}  \hspace{-4mm} &
\includegraphics[width=0.14\linewidth]{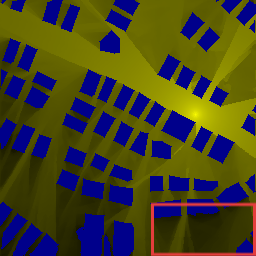}  \hspace{-4mm} &
\includegraphics[width=0.14\linewidth]{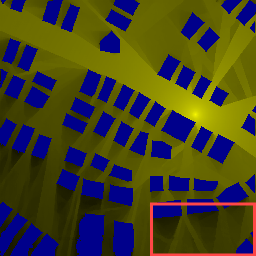}  \hspace{-4mm} &
\includegraphics[width=0.14\linewidth]{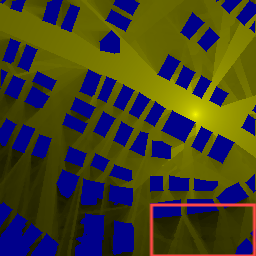}  \hspace{-4mm} &
\end{tabular}
\end{adjustbox}
\vspace{0.1mm}

\begin{adjustbox}{valign=t}
\begin{tabular}{ccccccccc}
\includegraphics[width=0.14\linewidth]{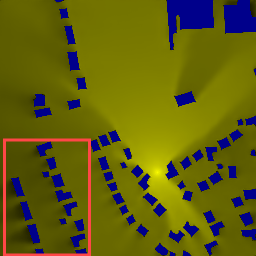} \hspace{-4mm} &
\includegraphics[width=0.14\linewidth]{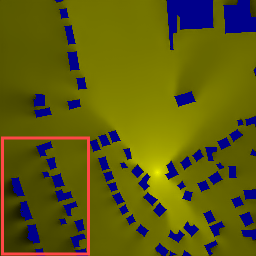}   \hspace{-4mm} &
\includegraphics[width=0.14\linewidth]{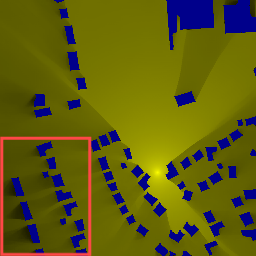}  \hspace{-4mm} &
\includegraphics[width=0.14\linewidth]{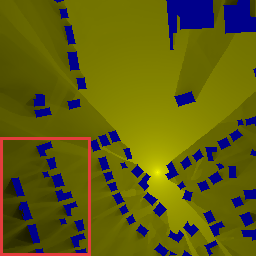}  \hspace{-4mm} &
\includegraphics[width=0.14\linewidth]{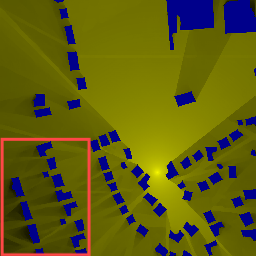}  \hspace{-4mm} &
\includegraphics[width=0.14\linewidth]{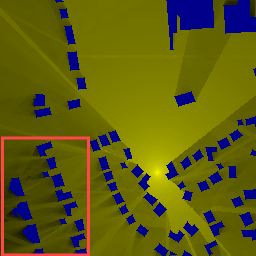}  \hspace{-4mm} &

\\
RME-GAN \hspace{-4mm} &
RadioUNet \hspace{-4mm} &
RadioDiff \hspace{-4mm} &
RMDM \hspace{-4mm} &
RadioDiff-$k^2$ \hspace{-4mm} &
Ground Truth
\\

\end{tabular}
\end{adjustbox}
\vspace{-10pt}
\caption{The comparisons of constructed MRM on different methods.}
\vspace{-12pt}
\label{fig-mrm}
% \vspace{0.1mm}
\end{figure*}

\subsection{Comparisons with SOTA Methods}
To provide a fair and informative benchmark, we compare iRadioDiff against four representative deep models that span the major architectural families; all baselines are trained and evaluated under the same data and experimental settings.
\begin{itemize}
    \item \textbf{RadioUNet} \cite{levie2021RadioUNet}: A widely adopted sampling-free convolutional neural network (CNN)-based method for RM reconstruction. It leverages a U-Net architecture trained with supervised learning to infer RMs directly from environmental information. Due to its simplicity and effectiveness, RadioUNet has become a foundational benchmark in RM construction research.
    \item \textbf{RME-GAN} \cite{zhang2023rme}: A generative adversarial network (GAN)-based model that originally incorporates both environmental features and sparse pathloss measurements (SPM) for RM generation. To ensure a fair comparison under the sampling-free setting, we adapt RME-GAN in our experiments to utilize only environmental features. While RME-GAN demonstrates the potential of adversarial training in generative RM tasks, it is not considered SOTA due to its reliance on sampled measurements in its original form.
    \item \textbf{RadioDiff} \cite{wang2024radiodiff}: The current state-of-the-art in sampling-free RM construction. RadioDiff formulates the task as a conditional generative problem based on a DDM. It combines a VAE and a denoising UNet to model reverse-time EM propagation dynamics in latent space. This architecture enables fine-grained reconstruction of pathloss textures and structural features, achieving superior performance in accuracy and perceptual quality.
    \item \textbf{RMDM} \cite{jia2025rmdm}: A conditional diffusion framework that synthesizes high-resolution RM in the image domain from environmental geometry and base-station metadata. Training incorporates a Helmholtz-equation PDE residual as a physics-alignment loss, fusing probabilistic generative priors with electromagnetic consistency for RM reconstruction.
\end{itemize}
Since RadioDiff employs the same backbone architecture as our proposed method and directly generates the RM from environmental and base station inputs, it effectively serves as an ablation baseline corresponding to the case without the map-outline guidance derived from the $k^2$-based analysis.

\subsection{Evaluation on DPM}
Based on the quantitative results in Table~\ref{tab-dpm} and the qualitative comparisons in Fig.~\ref{fig-srm} and Fig.~\ref{fig-drm}, the superiority of the proposed RadioDiff-$k^2$ method is clearly demonstrated across all evaluation metrics and RM types. Our method achieves substantial improvements over baseline approaches, particularly in its ability to capture fine-grained structural details and abrupt spatial variations caused by EM singularities. In the case of SRM, RadioDiff-$k^2$ reduces NMSE by over 40\% compared to RME-GAN and outperforms the strong diffusion-based baseline RadioDiff by a margin of 43\%. It also achieves the highest SSIM score of 0.9773, reflecting superior structural similarity, and the highest PSNR of 34.46 dB, indicating enhanced fidelity in reconstructing signal boundaries and texture features. For DRM, which are more challenging due to additional scattering and dynamic obstacles, RadioDiff-$k^2$ again leads with an NMSE of 0.0054, a 40\% reduction from the best-performing baseline, and shows consistent gains in RMSE, SSIM, and PSNR. This indicates that our method generalizes effectively to environments with more complex propagation characteristics.

The visual comparisons further support these conclusions. As shown in Fig.~\ref{fig-srm} and Fig.~\ref{fig-drm}, the RMs generated by RadioDiff-$k^2$ display significantly sharper textures, clearer signal gradients, and finer spatial transitions compared to other methods. Unlike baseline models, which often blur high-frequency structures and fail to localize abrupt changes, our approach precisely reconstructs EM singularities—regions corresponding to rapid pathloss changes induced by multipath and diffraction effects. These detailed reconstructions are crucial for real-world applications such as UAV trajectory planning and low-pilot CSI estimation, where local signal variation must be modeled accurately. In summary, both the numerical metrics and visual results validate that RadioDiff-$k^2$ achieves state-of-the-art performance in RM reconstruction, with particularly strong advantages in modeling complex multipath and singularity-dominated scenarios.

\subsection{Evaluation on IRT}
Based on the results shown in Fig.~\ref{fig-mrm} and the accompanying performance in Table~\ref{tab:irt}, our proposed RadioDiff-$k^2$ method demonstrates clear superiority over all baselines, particularly in the challenging MRM reconstruction scenario modeled using IRT4. While IRT-based datasets reflect the complex multipath propagation effects typically encountered in real-world wireless environments, they also pose significant challenges for accurate reconstruction, especially in modeling abrupt pathloss variations resulting from EM singularities. Quantitatively, RadioDiff-$k^2$ achieves an NMSE of 0.0066, reducing the error by 45.5\% compared to the best-performing baseline and significantly outperforming the standard RadioDiff. It also attains the lowest RMSE of 0.0236, and the highest SSIM of 0.9674 and PSNR of 32.68 dB scores, reflecting exceptional performance in preserving spatial structure and edge fidelity. These gains are critical in capturing high-frequency variations in EM intensity that result from reflection, diffraction, and scattering phenomena directly linked to the distribution of EM singularities.

Visually, as shown in Fig.~\ref{fig-mrm}, RadioDiff-$k^2$ produces significantly sharper RM images with clearer boundaries, richer textures, and finer detail compared to existing methods. Competing baselines tend to oversmooth or blur critical signal transitions, failing to capture the multipath-induced fluctuations observable in the ground truth. In contrast, our method successfully preserves localized high-gradient regions, making it particularly well-suited for applications like trajectory planning and beamforming in dense and dynamic 6G environments. Overall, the results validate the effectiveness of integrating physics-based EM singularity modeling into a generative diffusion framework, establishing RadioDiff-$k^2$ as a robust and high-fidelity solution for realistic, multipath-aware RM construction.

\subsection{Localization Performance Comparison}
To evaluate the practical utility of the generated radio maps, we performed localization experiments using a K-Nearest Neighbors (KNN) \cite{832252} algorithm with $K = 5$, randomly selecting 3,000 test points per map. These test positions include both Line-of-Sight (LoS) and Non-Line-of-Sight (NLoS) conditions, thereby capturing realistic localization challenges. As shown in Table~\ref{tab-localization}, across all three evaluation SRM, DRM, and MRM, RadioDiff-$k^2$ consistently achieves the lowest localization error. In particular, the average localization error remains below 5 meters in all cases, significantly outperforming prior methods such as RadioDiff, RMDM, and RadioUNet. These results demonstrate that the physics-informed singularity-aware structure of RadioDiff-$k^2$ enhances not only radio map reconstruction quality but also downstream localization accuracy under complex propagation conditions.
\begin{table}[h]
\captionsetup{font={small}, skip=16pt}
\centering
\caption{Localization error comparison on different methods in meters.}
\vspace{-10pt}
\resizebox{0.7\linewidth}{!}{
\begin{tabular}{@{}c|ccc@{}}
\toprule
Method & SRM & DRM & MRM \\ \midrule
RadioDiff-$k^2$    & {\color[HTML]{9A0000} \textbf{3.32}} & {\color[HTML]{9A0000} \textbf{3.87}} & {\color[HTML]{9A0000} \textbf{4.72}}  \\
RadioDiff          & {\color[HTML]{00009B} \underline{5.90}} & {\color[HTML]{00009B} \underline{7.18}} & {\color[HTML]{00009B} \underline{8.87}}  \\
RMDM               & 7.10 & 7.63 & 9.70  \\
RadioUNet          & 8.04 & 9.86 & 11.80  \\
RME-GAN            & 8.53 & 10.53 & 11.04 \\ \bottomrule
\end{tabular}
}
\label{tab-localization}
\end{table}

\subsection{Efficiency Comparison}
We evaluated the inference latency and memory consumption of all baseline and proposed models on an NVIDIA RTX Pro 6000 GPU. As shown in Table~\ref{tab-inference}, traditional models such as RME-GAN and RadioUNet exhibit extremely low inference times and modest memory footprints, owing to their relatively shallow architectures. In contrast, models based on diffusion processes, including RadioDiff, RMDM, and RadioDiff-$k^2$, incur approximately 3 to 4$\times$ higher memory usage due to the iterative nature and larger model capacities inherent in diffusion-based frameworks. Nevertheless, both RadioDiff and RadioDiff-$k^2$ leverage the LDM architecture, allowing them to maintain inference times under 1 second—significantly faster than RMDM's 19.3 seconds. This demonstrates that even for more realistic dynamic RM scenarios, such as those captured by DRM with moving vehicles and material heterogeneity, the inference latency of RadioDiff-$k^2$ remains practically acceptable for near real-time deployment.

\begin{table}[t]
\captionsetup{font={small}, skip=16pt}
\centering
\caption{Inference time and memory comparison of different models.}
\vspace{-10pt}
\resizebox{1\linewidth}{!}{
\begin{tabular}{@{}c|ccccc@{}}
\toprule
Model & RME-GAN & RadioUNet & RadioDiff & RMDM & RadioDiff-$k^2$\\ \midrule
Time (s) & 0.0057 & 0.0056 & 0.2400 &19.30 &0.7600 \\
Memory (GB) & 0.819 & 0.865 &1.520 &1.653 &3.020\\ \bottomrule
% Model & Time (s) & Memory (GB) \\ \midrule
% RME-GAN      & 0.0028 & 0.819 \\
% RadioUNet   & 0.0027 & 0.865 \\
% RadioDiff   & 0.2400 & 1.520  \\
% RMDM        & 19.030  & 1.653 \\
% RadioDiff-$k^2$ & 0.7600 & 3.020  \\ \bottomrule
\end{tabular}
}
\vspace{-10pt}
\label{tab-inference}
\end{table}

\subsection{Ablation Study}
\begin{table}[t]
\centering
\caption{Performance comparison using different outline maps as input for RadioDiff-$k^2$.}
\resizebox{0.90\linewidth}{!}{
\begin{tabular}{@{}c|c|cccc@{}}
\toprule
Method & Dataset & NMSE & RMSE & SSIM & PSNR \\ \midrule
\multirow{3}{*}{Canny} 
 & SRM    & 0.0414 & 0.0596 & 0.8831 & 24.85 \\
 & DRM & 0.0316 & 0.0495 & 0.8949 & 26.62 \\
 & MRM   & 0.6288 & 0.2614 & 0.3090 & 11.97 \\ \midrule
\multirow{3}{*}{LBP} 
 & SRM    & 0.8865 & 0.3096 & 0.1544 & 10.34 \\
 & DRM & 0.7529 & 0.2767 & 0.2448 & 11.38 \\
 & MRM   & 0.6297 & 0.2530 & 0.3085 & 12.06 \\ \bottomrule
\end{tabular}
}
\vspace{-12pt}
\label{tab-canny-lbp}
\end{table}
To further assess the importance of physically informed guidance, we replaced the Helmholtz-derived map outline in our framework with two classical texture extraction methods, Canny and LBP, and used them as condition inputs to the diffusion model. As shown in Table V, both alternatives result in significantly degraded performance across all datasets, particularly in terms of NMSE, SSIM, and PSNR. Compared to the results in Table~\ref{tab-canny-lbp} to Tables~\ref{tab-dpm} and Table~\ref{tab:irt}, the Canny and LBP based variants underperform the baseline RadioDiff model, which uses no outline guidance at all. This sharp contrast underscores the importance of using a physics-grounded indicator like $k^2_{\text{eff}}$ to extract meaningful singularity structures. Unlike heuristic edge detectors, the Helmholtz-informed outline preserves EM-consistent transitions, leading to more effective conditioning and improved radio map reconstruction.

\begin{figure*}[t]
\captionsetup{font={small}, skip=16pt}
    \centering
    % 第一行：SRM
    \subfigure[SRM]{
       \includegraphics[width=0.20\linewidth]{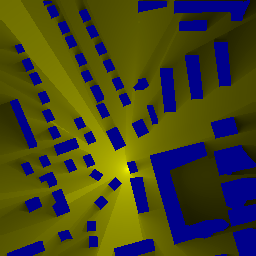}
       \includegraphics[width=0.20\linewidth]{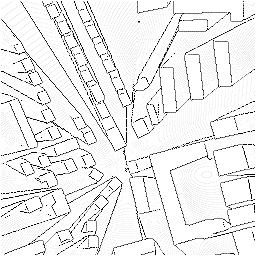}
       \includegraphics[width=0.20\linewidth]{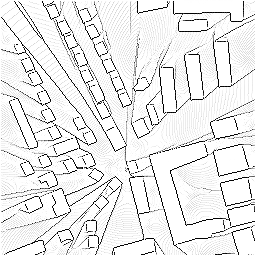}
       \includegraphics[width=0.20\linewidth]{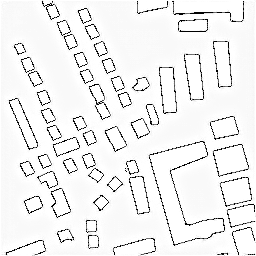}
    }
    
    % 第二行：DRM
    \subfigure[DRM]{
       \includegraphics[width=0.20\linewidth]{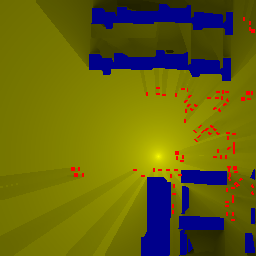}
       \includegraphics[width=0.20\linewidth]{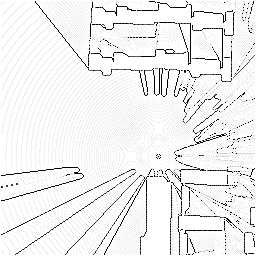}
       \includegraphics[width=0.20\linewidth]{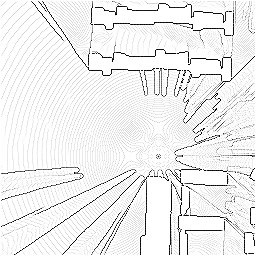}
       \includegraphics[width=0.20\linewidth]{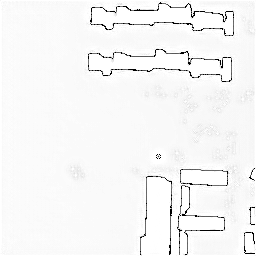}
    }
    
    % 第三行：IRT
    \subfigure[MRM]{
       \includegraphics[width=0.20\linewidth]{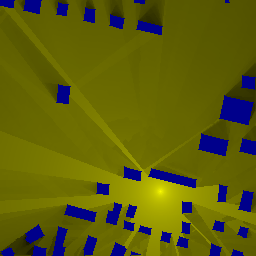}
       \includegraphics[width=0.20\linewidth]{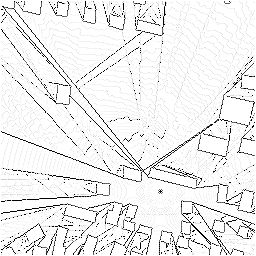}
       \includegraphics[width=0.20\linewidth]{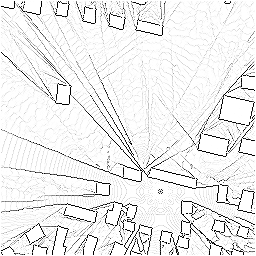}
       \includegraphics[width=0.20\linewidth]{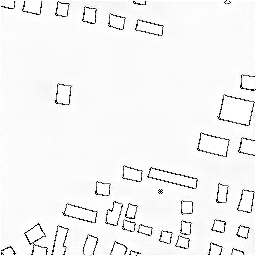}
    }

    \caption{Qualitative comparison of $k^2$ maps. The first column shows the ground-truth RM, the second column shows the ground-truth $k^2$ map derived from the RM, the third column presents the $k^2$ map predicted by the DM, and the fourth column displays the $k^2$ map predicted by a standard U-Net.}
    \label{fig:qualitative_comparison}
\end{figure*}

\begin{table}[t]
\centering
\caption{Quantitativ Comparison on $k^2$ map.}
\resizebox{0.9\linewidth}{!}{
\begin{tabular}{@{}c|c|c|c|c@{}}
\toprule
Metric & Model  & SRM & DRM & MRM \\ \midrule
\multirow{2}{*}{DTC$\uparrow$} 
 & UNet       & 0.8141 & 0.6857 & 0.7433 \\
 & Diffusion  & 0.9703 & 0.9782 & 0.9577 \\\midrule
\multirow{2}{*}{DTIoU$\uparrow$} 
 & UNet       & 0.8283 & 0.6997 & 0.7483 \\
 & Diffusion  & 0.9023 & 0.8403 & 0.8615 \\ \midrule
\multirow{2}{*}{LPIPS} 
 & UNet       & 0.2042 & 0.2449 & 0.2590 \\
 & Diffusion  & 0.0933 & 0.1107 & 0.1313 \\ \bottomrule
\end{tabular}
}
\vspace{-6pt}
\label{tab:contour_metrics}
\end{table}

To investigate the impact of network architecture on the accuracy of $k^2$ map prediction, we compare a standard U-Net with our proposed diffusion-based model across three RM datasets. As shown in Table~\ref{tab:contour_metrics}, the diffusion model significantly outperforms the U-Net across all evaluation metrics, including the distance-tolerant coverage (DTC), distance-tolerant intersection over union (DTIoU), and learned perceptual image patch similarity (LPIPS) \cite{zhang2018unreasonable}. These improvements underscore the superior capacity of the diffusion backbone in capturing the fine-grained singularity structures that are often missed by conventional architectures. Furthermore, the qualitative results in Fig.\ref{fig:qualitative_comparison} provide intuitive visual evidence of this performance gap. The $k^2$ maps generated by the diffusion model closely resemble the ground-truth singularity patterns and exhibit sharper texture boundaries, effectively highlighting abrupt transitions in the radio field. In contrast, the U-Net predictions are severely smoothed, lacking clear structural detail, which aligns with the blurred edges observed in its RM outputs in Fig.~\ref{fig-srm} to Fig.~\ref{fig-mrm}. These results demonstrate that the use of diffusion models not only benefits full RM reconstruction but also enhances the fidelity of intermediate physics-informed guidance signals such as the $k^2$ map.

\section{Conclusion}
In this work, we have proposed RadioDiff-\(k^2\), a physics-informed, diffusion-based framework for multipath-aware RM construction. By incorporating a Helmholtz-informed curvature indicator, the method first localizes electromagnetic singularities and then conditions a second-stage generative model on these features, improving interpretability and reconstruction fidelity. Extensive experiments on static and dynamic multipath environments show substantial gains over state-of-the-art baselines. The approach is lightweight at inference and has immediate applicability to real-time wireless tasks such as UAV trajectory planning and IRS beam selection; future work will extend robustness to material heterogeneity via conditional dielectric estimates. In future work, regarding material-mismatch realism, large-scale datasets with per-object heterogeneous dielectric/conductive annotations are currently unavailable; we will incorporate estimated environmental electromagnetic parameters as additional conditional inputs to the diffusion model to improve robustness to material heterogeneity scenarios.

\bibliography{ref}
\bibliographystyle{IEEEtran}
\ifCLASSOPTIONcaptionsoff
  \newpage
\fi
\end{document}